\documentclass[conference]{IEEEtran}
\IEEEoverridecommandlockouts
\usepackage{cite}
\usepackage{amsmath,amssymb,amsfonts}
\usepackage{algorithmic}
\usepackage{graphicx}
\usepackage{textcomp}
\usepackage{xcolor}
\usepackage[caption=true, font=footnotesize, labelfont=sf, textfont=sf]{subfig}

\newcommand{\bbE}{\mathbb{E}}

\newcommand{\etal}{\textit{et al.}}
\def\BibTeX{{\rm B\kern-.05em{\sc i\kern-.025em b}\kern-.08em
    T\kern-.1667em\lower.7ex\hbox{E}\kern-.125emX}}

\begin{document}

\title{Modeling Lost Information in Lossy Image Compression\\
\thanks{\textsuperscript{*}Note: Work done during an internship at Microsoft Research Asia.}
}

\author{\IEEEauthorblockN{Yaolong Wang*}
\IEEEauthorblockA{ \textit{University of Toronto}\\
Toronto, Canada \\
yaolong.wang@mail.utoronto.ca}
\\
\IEEEauthorblockN{Shuxin Zheng}
\IEEEauthorblockA{\textit{Microsoft Research Asia}\\
Beijing, China \\
shuz@microsoft.com}
\and
\IEEEauthorblockN{Mingqing Xiao*}
\IEEEauthorblockA{\textit{Peking University}\\
Beijing, China \\
mingqing\_xiao@pku.edu.cn}
\and
\IEEEauthorblockN{Chang Liu}
\IEEEauthorblockA{\textit{Microsoft Research Asia}\\
Beijing, China \\
changliu@microsoft.com}
\\
\IEEEauthorblockN{Tie-Yan Liu}
\IEEEauthorblockA{\textit{Microsoft Research Asia}\\
Beijing, China \\
tyliu@microsoft.com}
}

\maketitle

\begin{abstract}
Lossy image compression is one of the most commonly used operators for digital images. Most recently proposed deep-learning-based image compression methods leverage the auto-encoder structure, and reach a series of promising results in this field. The images are encoded into low dimensional latent features first, and entropy coded subsequently by exploiting the statistical redundancy. However, the information lost during encoding is unfortunately inevitable, which poses a significant challenge to the decoder to reconstruct the original images. In this work, we propose a novel invertible framework called Invertible Lossy Compression (ILC) to largely mitigate the information loss problem. Specifically, ILC introduces an invertible encoding module to replace the encoder-decoder structure to produce the low dimensional informative latent representation, meanwhile, transform the lost information into an auxiliary latent variable that won't be further coded or stored. The latent representation is quantized and encoded into bit-stream, and the latent variable is forced to follow a specified distribution, i.e. isotropic Gaussian distribution. In this way, recovering the original image is made tractable by easily drawing a surrogate latent variable and applying the inverse pass of the module with the sampled variable and decoded latent features. Experimental results demonstrate that with a new component replacing the auto-encoder in image compression methods, ILC can significantly outperform the baseline method on extensive benchmark datasets by combining with the existing compression algorithms.

\end{abstract}


\section{Introduction}
Lossy image compression has played a central role in the scenario of image storing and transferring for a long time, especially given the exploding amount of large-sized images and comparatively limited storage or bandwidth nowadays. Specifically, transform coding based methods perform well and are widely adopted in practice, such as JPEG~\cite{jpeg}, JPEG2000~\cite{jpeg2000} and BPG~\cite{bpg}. Recently, deep-learning-based lossy image compression methods~\cite{balle1,balle2,balle3,liu2019non,balle4,van2020lossy,agustsson2019generative,balle5,tschannen2018deep,ma2019image} have generated great interests due to the impressing performance and low bitrate. 

Unlike previous transform coding methods, deep-learning-based methods first transform an image $x$ into a lower-dimensional latent representation vector $y$ by an encoder network, then quantize $y$ as a discrete-valued vector $\hat{y}$. Lossless entropy coding methods, such as arithmetic coding~\cite{rissanen1981universal}, are then applied to $\hat{y}$ and compress it into a bit-stream. Some prior works~\cite{balle1,balle2,balle3,lee2019extended,liu2019non} adopt an auxiliary network as an entropy model to estimate the density and provide the statistics to the entropy model. A decoder network is used to approximate the inverse function that maps the latent variables back to pixels. There inevitably exists information loss after passing the encoder network by reducing the representation dimensions and quantization by rounding floating numbers.

The reduction in the dimensionality of the representation on $y$ induces a significant drop in the amount of information maintained by the original image $x$, which is inconsistent with the goal of compression that aims to reduce the entropy of the representation under a prior probability entropy model. Previous works~\cite{balle2,balle3} mainly focus on finding a reasonable estimation of the prior density to minimize the expected length of the bit-stream, but few put effort into improving the expected distortion of the reconstructed image with respect to the original $x$. The information loss poses a notable challenge to recovering the original image $x$ by using a decoder network only since the task is made ill-posed, which would obviously influence the rate-distortion optimization. One way to keep all the information during encoding is to preserve the dimensionality of the representation in $y$, i.e. the same as the original image $x$. The reconstruction stage would benefit from such an informative representation, while it is extremely tough to encode such high-dimensional data into desired low bitrate with lossless entropy coding methods.

In this paper, we propose a novel framework for lossy image compression called Invertible Lossy Compression (ILC) to tackle this intractable problem by capturing the most knowledge of the lost information. The recovery of the original image $x$ can benefit from modelling the statistical behaviour of the lost information in expectation during encoding. To this end, the lost information should be expressed in an explicit form and can be encoded to hidden states without unconscious information loss. The encoder-decoder framework employed in previous image compression methods does not well meet such requirement since massive efforts have to be paid to make the encoder invertible and the decoder as the inverse of it, and even so there is still approximation error in the invertibility. Therefore, we adopt an invertible encoding module (IEM), which is strictly reversible and apparently be more satiable to this pair of inverse tasks. IEM consists of two essential components, i.e., invertible downsampling layers to enlarge the receptive field and decompose the spatial correlation among channels; and coupling layer to enhance the expressive power of the module. Since an invertible model gives a representation of the same dimensionality with input image $x$, we split the encoded representation as $y$ and $z$, where $y$ preserves the necessary information for image reconstruction and $z$ stores all left information (like high-frequency noise that doesn't alter the original image much). For lossy compression, we discard $z$ and encode the quantized $\hat{y}$ into a bit-stream, while we also model the distribution of the lost information $z$. For reconstruction, we feed the INN inversely with $\hat{y}$ and a randomly drawn sample of $z$ from the learned distribution.

To achieve our purpose, as ensuring the representation of $y$ to be informative as well as coding favourable, it is crucial to make IEM capturing the knowledge in the distribution of $x$, and meanwhile removing the dependency between $y$ and $z$. To this end, we employ a distribution matching loss to encourage $z$ following a Gaussian distribution. More importantly, the quantization process would change the distribution on the representation from $y$ to $\hat{y}$, which disturbs the inverse procedure of IEM when recovering $x$ by combining with $z$. It is challenging to train an invertible model with such a distribution mismatch. Observing that conventional model training is more robust to the quantization process, we propose a knowledge distillation module (KDM) to transfer the knowledge from encoder-decoder models to IEM with knowledge distillation~\cite{bucilua2006model,ba2014deep,hinton2015distilling}. It usually starts by training a teacher model, an encoder-decoder compression model in our task, and then optimize the target, invertible student model, so that it mimics the teacher model's behaviours. Empirically, we choose the output of the encoder network as the objective for distillation, combining reconstruction loss, entropy loss, and distribution loss to optimize efficiency. The KDM would provide soft labels and accelerate the convergence of invertible models to match the teacher encoder-decoder model's performance quickly. We conduct experiments on extensive benchmark datasets, and the comparison with the baseline method has shown significant improvement that proves the effectiveness of our proposed ILC.

The main contributions of this paper are highlighted as follows:
\begin{enumerate}
    \item We propose a novel ILC framework by introducing an invertible encoding module into lossy image compression with an encoder-decoder framework to simultaneously produce low-dimensional informative representations and capture the knowledge in the distribution of the lost information during encoding.
    \item We propose an efficient training objective of IEM to encourage the latent variable to obey a pre-specified distribution, which can be easily drawn and provide rich information during decoding. We then employ a knowledge distillation module drawn from the teacher model to overcome the distribution mismatch problem brought by quantization in compression and accelerate the optimization of IEM.
    \item Extensive experimental results demonstrate that simply replacing the encoder and decoder models by our ILC framework can provide a performance boost of reconstructed image compression results on various benchmark datasets.
\end{enumerate}

\section{Related Work}

\subsection{Lossy Image Compression}
The traditional lossy compression algorithms for images~\cite{jpeg,jpeg2000,bpg} use prior knowledge to decouple the low- and high-frequency signals, (e.g. discrete cosine transformation (DCT), discrete wavelet transformation (DWT), etc.), and perform reversible coding algorithms (e.g. Huffman Coding~\cite{van1976construction}) on the quantized signals to achieve compression.

In recent years, more and more deep-learning-based works came into sight and attained astonishing results~\cite{balle1,balle2,balle3,liu2019non,balle4,van2020lossy,balle5,lee2019context}. Generally, they use auto-encoders, which is widely employed in representation learning and generative problems. The informative bottleneck representation in the auto-encoder framework is well-suited for lossy compression. It is trying to use a neural network, instead of prior knowledge, as an encoder, to draw a lower-dimensional representation from the image straightly, and quantize the low-frequency signal such that it can be coded into a bit-stream. When decompressing the image, a decoder network is employed to reconstruct the original image from the decoded quantized latent representation.

However, quantization leads to differentiability issues. For the past a few years, to enable the entire optimization process to be carried out end-to-end, it has reached a consensus that a uniform noise needs to be superimposed on low-frequency signals as a soft quantization during training. On the other hand, the previous works are also roughly the same in the choice of coding algorithm, that is, the use of arithmetic coding, an efficient coding algorithm based on entropy modelling. It then raises the question of how to estimate the entropy of the signal variable better. Much work on this issue has accomplished positive progress. For example, hyperprior and autoregressive components are used~\cite{balle2,balle3} to jointly model the latent feature map such that a better-performing entropy model can be constructed. Besides, there is also some work on auto-encoder, which helps improve the neural network's capacity to extract features. It introduced a non-local attention block to assist auto-encoder to capture the local and global correlation of the latent feature map~\cite{liu2019non}.

Although previous methods share promising performance, they ignore the difficulty in the reconstruction stage brought by non-negligible information loss during encoding. In this work, we identify the problem and propose to mitigate by explicitly modelling the lost information.

\begin{figure*} [!t]
    \centering
      \subfloat[]{
       \includegraphics[width=0.35\textwidth]{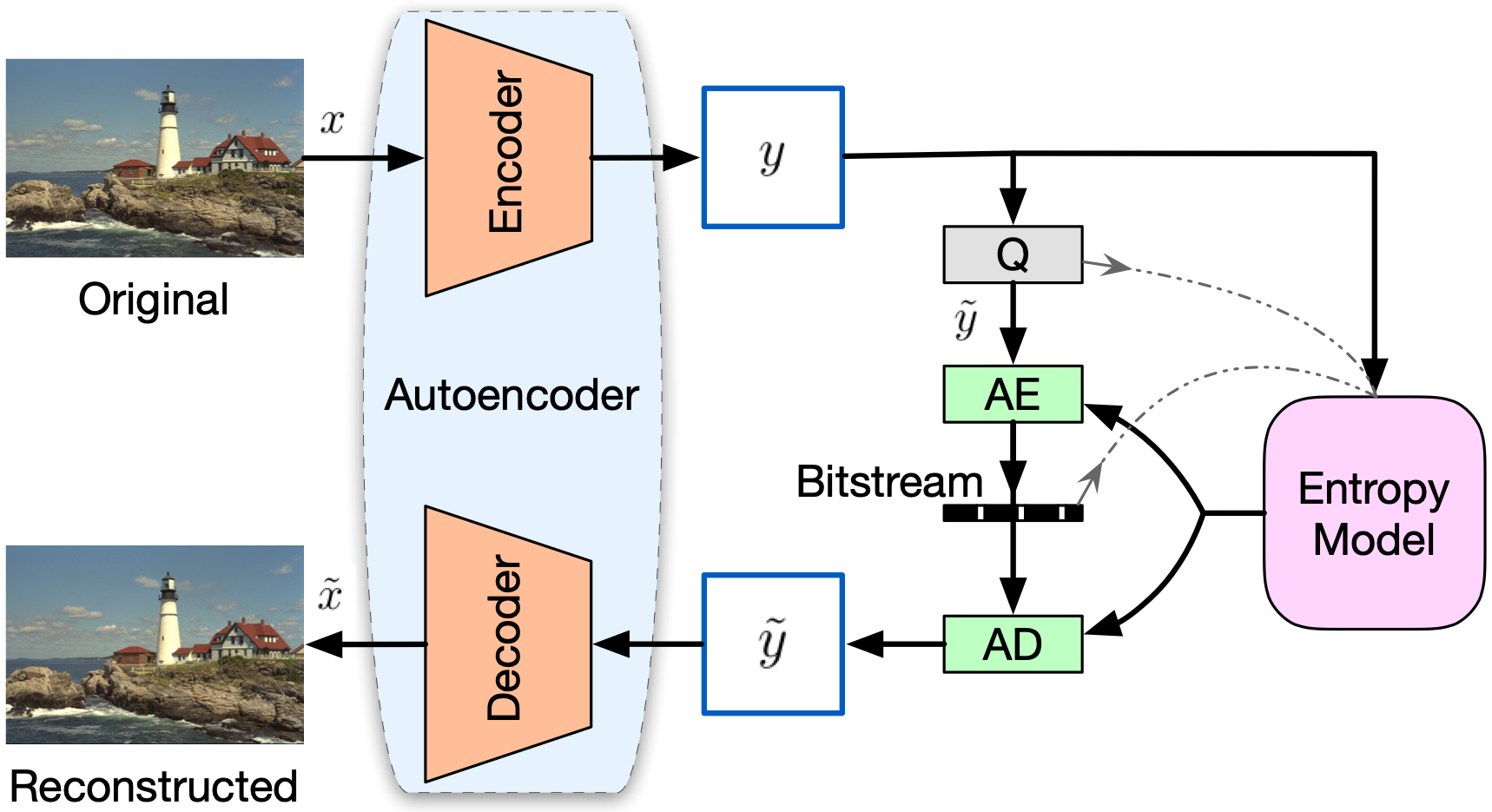}
    \label{fig.1a}}\hfill
      \subfloat[]{
        \includegraphics[width=0.55\textwidth]{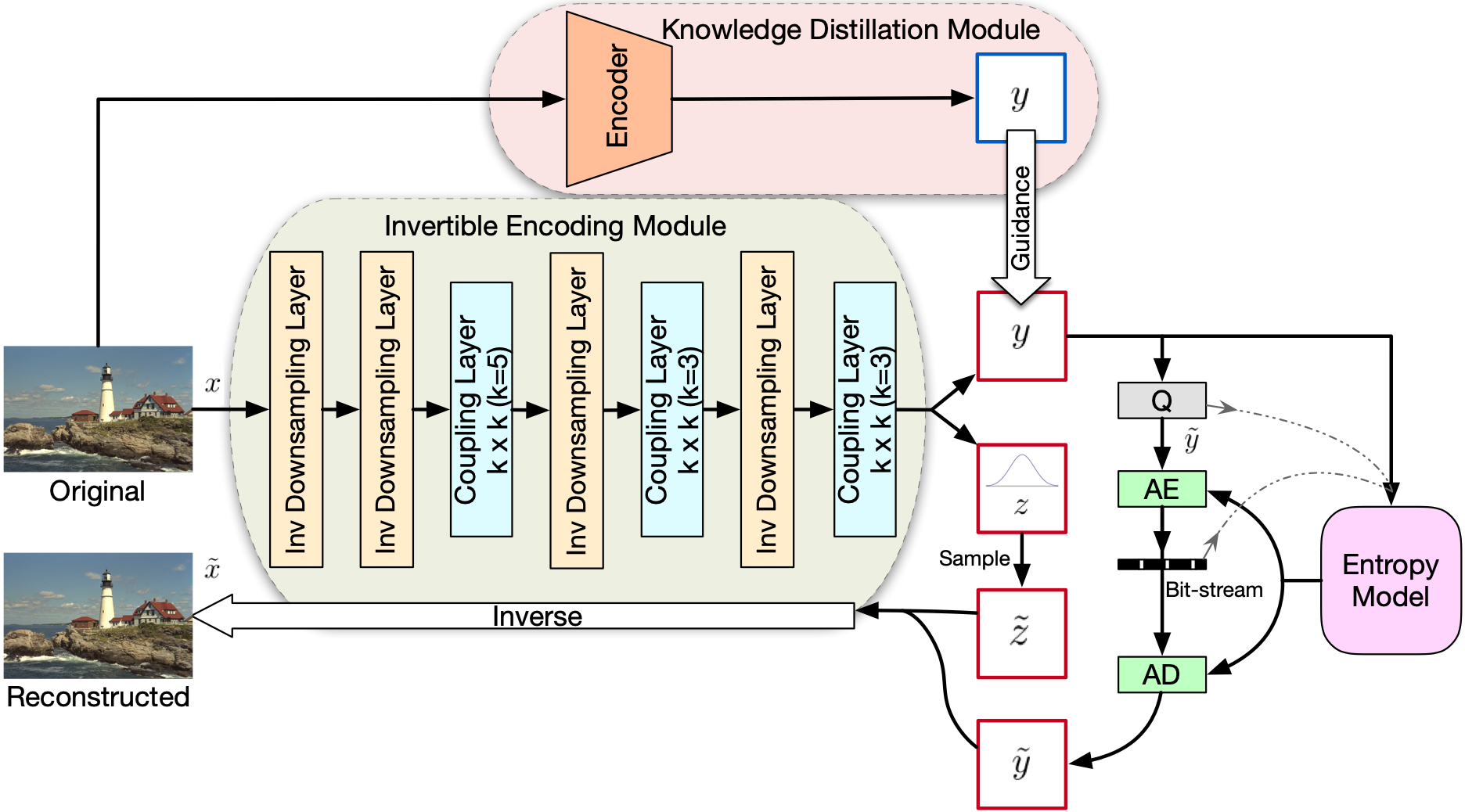}
    \label{fig.1b}}
      \caption{(a): the general frameworks using variational auto-encoder, where the entropy model generally refers to all the modules that can estimate the density function of y. It can be expanded to jointly estimate with hyperprior or autoregressive component~\cite{balle2,balle3}. (b): the general framework of ILC. An invertible encoding module takes the place of both encoder and decoder network in (a), and a knowledge distillation module is adopted during training stage.}
      \label{fig.1} 
\end{figure*}
\subsection{Invertible Neural Network}
It has been shown in lots of scenarios that the neural network with invertibility as its core design principle can achieve the same or even better performance than the non-invertible neural networks~\cite{kingma2018glow,xiao2020invertible,gomez2017reversible}.

As for generative models, denoting the input data as $x$, an invertible neural network as $f_{\theta}$, and $f_{\theta}(x)$ as $z$, the inverse function $f_{\theta}^{-1}$ can be trivially obtained, such that $P_X(x)$ can be easily sampled with $x=f_{\theta}^{-1}(z)$, where $z \sim P_Z(z)$. Furthermore, the density function of $P_Z(z)$ can be explicitly defined, which allows us to use the maximum likelihood method for training. Besides, GANs and VAEs are also well-known in generative problems, but both have defects. In VAEs, the distribution of latent variables, $P_Z(z)$, can only be approximately inferred by data, which means the entire training objective is not an exact form, but a variational lower bound on the log-likelihood of $P_X(x)$. And for GANs, due to the lack of encoder, it cannot perform inference, which significantly impedes the usability in our scenario. While, with invertibility, a neural network can not only accurately evaluate log-likelihood and perform inference, but be naturally superior in synthesis. 

In order to ensure the invertibility of the system, each sub-block of the neural network is designed to be invertible, which makes maintaining the model capacity become a top priority task while limiting the design of structure. To address this concern, we take the coupling layer with affine transformation, introduced in RealNVP~\cite{dinh2017density}, as a general solution. Consider the $l^{th}$ coupling layer. Given a $D$ dimensional input $h^l$, and a slicing position $d < D$, the $D$ dimensional output $h^{l+1}$ of an affine coupling layer follows the equations: 
\begin{eqnarray}\label{eq:1}
    \begin{aligned}
        &h^{l+1}_{1:d} = h^l_{1:d} \odot \exp(\alpha \cdot \psi(h^l_{d+1:D})) + \phi(h^l_{d+1:D}),\\
        &h^{l+1}_{d+1:D} = h^l_{d+1:D} \odot \exp(\alpha \cdot \rho(h^{l+1}_{1:d})) + \eta(h^{l+1}_{1:d}),
    \end{aligned}
\end{eqnarray}
where $\psi$, $\phi$, $\rho$ and $\eta$ are arbitrary dimensional invariant functions, and $\alpha$ is a constant factor served as a clamp.

For the inverse, given a $D$ dimensional $h^{l+1}$, and a slicing position $d < D$, the $D$ dimensional $h^{l}$ follows: 
\begin{eqnarray}\label{eq:2}
    \begin{aligned}
        &h^l_{d+1:D} = (h^{l+1}_{d+1:D} - \eta(h^{l+1}_{1:d})) \odot \exp(-\alpha \cdot \rho(h^{l+1}_{1:d})),\\
        &h^l_{1:d} = (h^{l+1}_{1:d} - \phi(h^l_{d+1:D})) \odot \exp(-\alpha \cdot \psi(h^l_{d+1:D})).
    \end{aligned}
\end{eqnarray}

INN has also been applied to paired data $(x,y)$ and this idea has been demonstrated in different problems. For example, Ardizzone \etal~\cite{ardizzone2019analyzing} analyzed real-world problems from medicine and astrophysics. In image compression tasks, the classical Maximum Mean Discrepancy (MMD)~\cite{dziugaite2015training} method fails to measure the difference in such high-dimensional probability distributions. A conditional INN~\cite{ardizzone2019guided} is designed is applied to guided image generation and colorization. In their task, the guidance $y$ is given as a condition that is obviously not suitable for our aim. Recently, Xiao \etal \cite{xiao2020invertible} propose to use INN as a transformation between high- and low- resolution images. In their task, no explicit constraint of entropy on the low-dimensional representation $y$ is considered, which would be one of the biggest challenges in image compression tasks.

\section{Invertible Lossy Compression}


\begin{figure*} 
    \centering
      \subfloat[]{
       \includegraphics[scale=0.23]{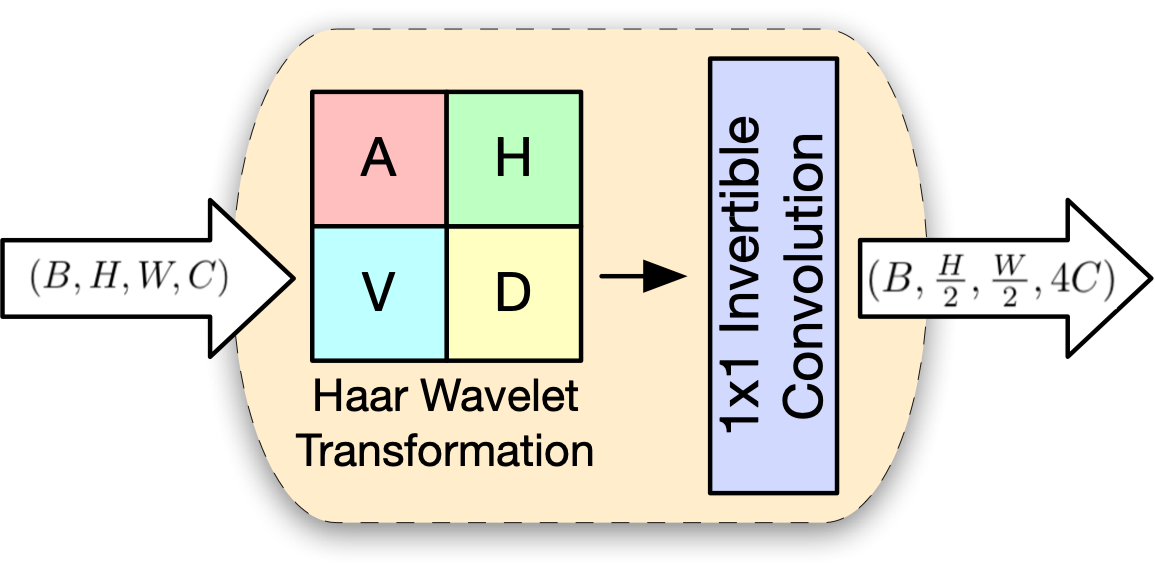}
    \label{fig.2a}}\hfill
      \subfloat[]{
        \includegraphics[scale=0.205]{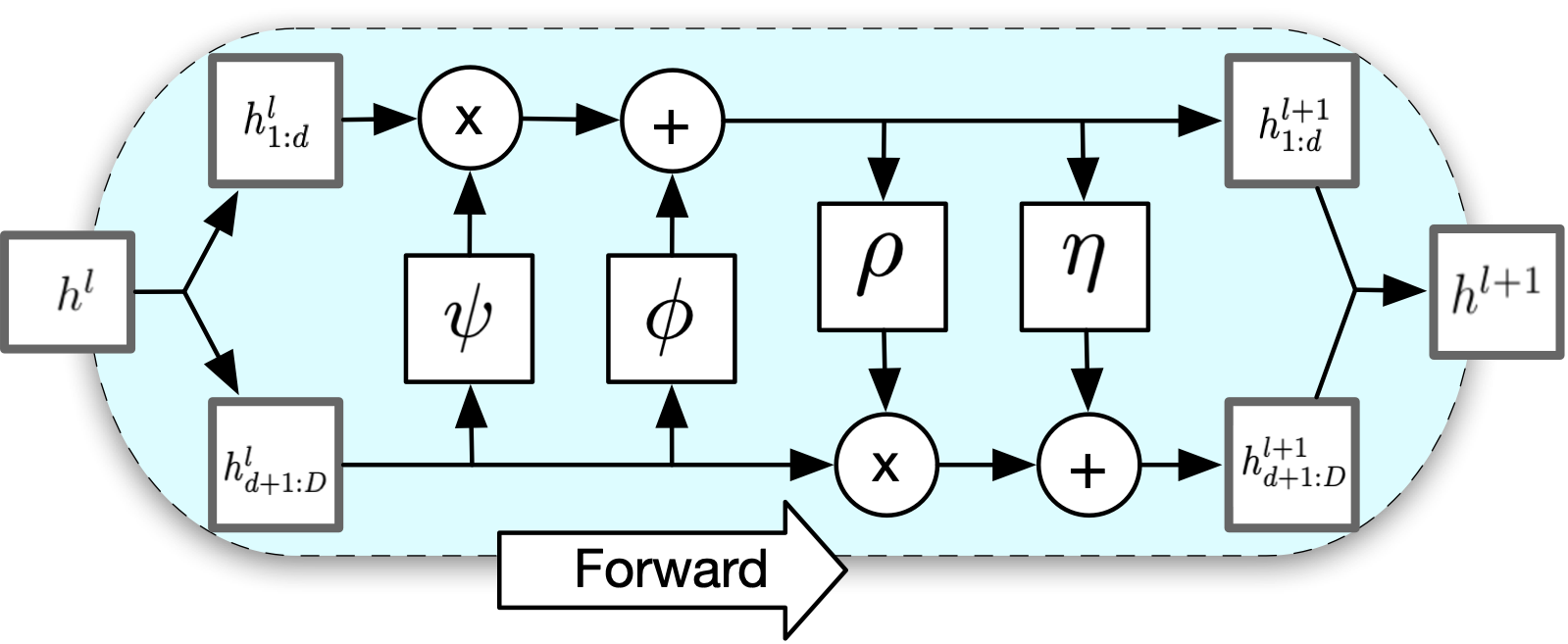}
    \label{fig.2b}}\hfill
      \subfloat[]{
        \includegraphics[scale=0.26]{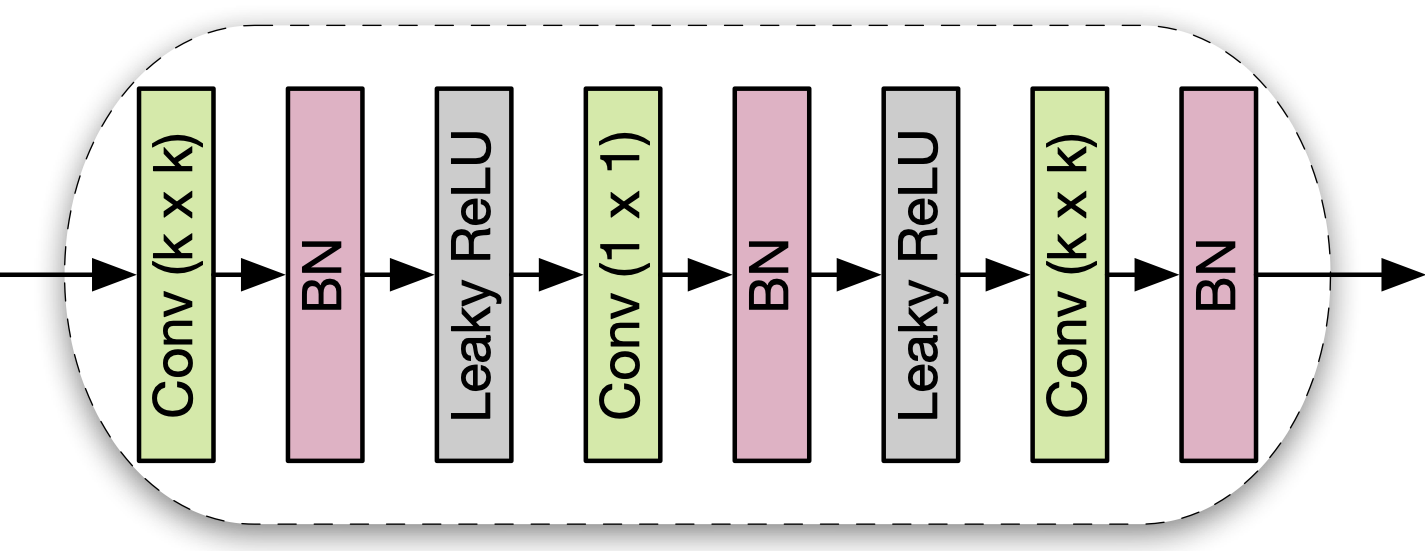}
    \label{fig.2c}}
      \caption{The architecture of IEM, where (a): the invertible downsampling layer, (b): computation graph for forward propagation in the coupling layer and (c): details of the operation function $\psi (\cdot), \phi(\cdot),\rho(\cdot)$ and $\eta(\cdot)$.}
      \label{fig.2} 
\end{figure*}


Figure \ref{fig.1a} shows the general frameworks of previous variational auto-encoder based lossy image compression~\cite{balle1,balle2,balle3}. Typically it consists of three modules: an encoder network to transform original image $x$ into a low-dimensional representation $y$, an entropy model to estimate the density of the quantized representation $\hat{y}$, and a decoder network to reconstruct the image from $\hat{y}$. Figure~\ref{fig.1b} illustrates the sketch of the general framework of ILC. The main difference between ILC and previous methods is that instead of a commonly used encoder-decoder module, an invertible encoding module (IEM) is adopted to explicitly model the lost information during encoding, which mitigates the difficulty in reconstruction stage. In the forward procedure, image $x$ is transformed into a coding target $y$ and an auxiliary latent variable $z$. We enforce the transformed $z$ to follow a specified distribution $p(z)$ independent of $y$ by a network, which captures the statistical knowledge of the lost information for reconstruction. This can be achieved since it is guaranteed that any absolutely continuous distribution can be transformed bijectively to a standard Gaussian~\cite{hyvarinen1999nonlinear} through the knowledge in the network. Then for the inverse, we can utilize a sample $z'\sim p(z)$ to replace the original $z$ and reconstruct image $x$ from quantized $\hat{y}$ through the inverse transformation. Furthermore, a knowledge distillation module (KDM) would be coupled during the training process to stable the optimization and accelerate the training convergence. We will describe them in detail as follows.

\subsection{Invertible Encoding Module}
Our IEM consists of two basic invertible components, i.e., invertible downsampling layer and coupling layer. As shown in Figure~\ref{fig.1}, IEM takes the place of both the encoder and decoder modules in our framework. When the image $x$ is being compressed, IEM acts s the encoder by taking $x$ as input and feeding into the forward direction of the model. In contrast, IEM is inversely applied to serve as the decoder when the compressed image is being reconstructed.

{\textbf{Invertible Downsampling Layer}}\quad
Downsampling is necessary for the entropy model to enlarge the receptive field and decompose the spatial correlation in compression. However, the traditional downsampling layer in previous learning-based compression methods is obviously irreversible. Therefore, we carefully design an invertible downsampling layer to achieve our purpose. Figure~\ref{fig.2a} shows the detailed architecture. Firstly, we introduce a wavelet transformation as one of our basic modules. The inverse function of each wavelet transformation naturally exists. More importantly, strong prior knowledge is provided by wavelet transformations that explicitly separate the low- and high- frequency signals, from which the image compression algorithm would definitely benefit~\cite{antonini1992image}. On the contrary, all previous deep-learning-based methods fail to integrate this into their models. Specifically speaking, the wavelet transformation transforms the input tensor with height $H$, width $W$, and channel $C$ into a tensor of shape $(\frac{1}{2}H, \frac{1}{2}W, 4C)$, where the first $C$ slices are low-frequency contents equivalent to average pooling, and the rest are high-frequency components corresponding to residuals in the vertical, horizontal and diagonal directions. We employ Haar wavelet transformation in our framework, which is easy to implement while enlarging the receptive field without information loss and containing certain prior knowledge.

Although the non-learnable wavelet transformation can downsample the input image to enlarge the receptive fields, provide strong prior for the model and be strictly invertible, it suffers from the fixed split of information, i.e. only the first quarter are main low-frequency contents. A more flexible separation can adapt our channel split between $y$ and $z$ better and makes training more accessible. Therefore, we use a 1$\times$1 invertible convolution after the wavelet transformation to refine the split of information. 1$\times$1 invertible convolution is initially proposed in GLOW~\cite{kingma2018glow}, which is originally used for the channel-wise permutation. Different from their purpose, we leverage it to make channel-wise refinement after the wavelet transformation and initialize its weight as an identity matrix.

{\textbf{Coupling Layer}}\quad 
Processed by invertible downsampling layer, the feature map is roughly broken down into two segments, a low-frequency component carrying the majority of information and a high-frequency component mainly composed of the former's residuals, and then fed into a stack of convolutional layers applied to further abstract the corresponding signals. Taking $h^l$ as an input of the $l^{th}$ coupling layer, the feature map is partitioned into two components by $1:2$, denoted as $h^l_{low}$ and $h^l_{high}$ respectively. Notice that the ratio is consistent with the proportion of dimensions in the final outputs, $y$ and $z$. To further decouple signals, we employ an affine transformation, proposed in RealNVP~\cite{dinh2017density}, on the high-frequency feature map and fuse the captured signal into $h^l_{low}$. On the other hand, if the compression rate is low, we expect simple information to be transferred into $h^l_{high}$ so that we can model it in the latent space, rather than eliminated in quantization. As a result, we apply another affine transformation for the flow of information from the low- to the high-frequency component. The detailed structure is presented in \ref{fig.2b}.

The complete expression of affine transformation is shown in Eq.\ref{eq:1}, where the constant factor, $\alpha$, is set to 1, and all the transformation functions (i.e., $\phi(\cdot), \psi(\cdot), \rho(\cdot)$ and $\eta(\cdot)$) can be arbitrary as long as the dimensionalities of input and output are matched. To enhance the expressive power of our model while retaining the light-weight computation, we employ a simple but effective bottleneck-like structure as the transformation functions, which is shown in Figure~\ref{fig.2c}.

\subsection{Knowledge Distillation Module}

Yet, there are still several challenges that may significantly influence the optimization process of IEM. On the one hand, due to the invertibility, IEM can transform the distribution of $x$ and $(y,z)$ to each other. However, the quantization on $y$ would apparently change the distribution on $(y,z)$, and we expect that IEM would fail to be robust to such data jitter prompted by quantization, and induce a non-negligible drop in performance. On the other hand, guaranteeing the invertibility, it has to make $y$ independent from $z$ as much as possible and force both of them to follow the distributions as required. Considering the various conditions, the massive inequality in the amount of information between $y$ and $z$ would further provoke a remarkable increment in the difficulty of optimization. 

Therefore, we introduce the KDM to mitigate these stubborn challenges. Observing that the encoder-decoder frameworks are much robust to this distribution mismatch caused by the quantization process, which is most likely due to the separate training procedure, we propose to use the encoder network as a teacher model and encourage IEM to mimic the output representation of it at the training stage. In this way, IEM can learn to output the representation with less distribution changing during the quantization process, which makes the inverse function of IEM more robust to the distribution mismatch than before. More importantly, KDM can obviously increase the utility and the ease of training in the early stage, such that IEM can refine the distribution of $y$ from the old encoder, and surrounding it, look for a better low-dimensional manifold of the data distribution that can be efficiently inverted. 

Practically, it requires a fully specified prior, a teacher encoder and its affiliated entropy module, which could furnish at least a sub-optimal for the sophisticated knowledge in the distribution on $y$ with learnable uncertain components. Specifically speaking, the fully trained prior is utilized to initialize the entropy model for IEM. Throughout the training, the teacher's knowledge is transferred through a distillation loss guiding $y$, which is evaluated by first feeding the same $x$ into both the teacher encoder and IEM and then computing their difference on $y$. 

\subsection{Optimization Objective}

Our fundamental training objective follows previous learning-based lossy compression methods, i.e. to minimize a weighted sum of the rate and distortion $L = R + \lambda D$, where the rate $R$ is lower-bounded by the entropy of the discrete probability distribution of the quantized vector $\mathbb{H}[P_q]$, and the distortion $D$ is the difference between the reconstructed $\hat{x}$ and the original input $x$. In addition, two novel objectives are included for our invertible framework and efficient training: (1) A distribution matching loss for capturing the distribution of lost information, and (2) A distillation loss that stables our training procedure.

\textbf{Rate}\quad
Our basic goal is to minimize the rate of the coding target $y$. Therefore, we leverage entropy as our objective, which is consistent with previous works~\cite{balle1,balle2,balle3}. As mentioned above, an entropy module is used to estimate the entropy on $y$. We follow their loss definition and denote it as: 
\begin{align}
	L_{\mathrm{rate}}(\theta) := \sum_{n}^{N} -\mathbb{E}[\log_2 P_{\tilde{y}}(f^y_{\theta}(x^{(n)}) + \Delta y)], 
	\label{eq:rate}
\end{align}
where $\tilde{y} = f^y_{\theta}(x^{(n)}) + \Delta y$ is the approximation of the quantizaiton and $\Delta y$ is an additive i.i.d. uniform noise, with the same with as quantization bins, which in our case is one.

\textbf{Distortion}\quad
Because of the uncertainty of $z$, and the quantization on $y$, a distortion loss is still required to ensure the inverse reconstruction of our model. We denote the reverse process as $f_{\theta}^{-1}(\hat{y}, z)$. The distortion loss is able to encourage our model to adapt the new sample $z$ drawn from $p(z)$. We formulate the loss as:

\begin{align}
	& L_{\mathrm{distortion}}(\theta) := \sum_{n}^{N} \ell_x(f^{-1}_{\theta}(\hat{y}, z), x^{(n)}),
	\label{eq:distortion}
\end{align}

where $z$ is a sample from $p(z)$. For training stability in practice, we empirically take the most-likely sample from the distribution for reconstruction. We employ $L_2$ loss as $l_x(\cdot)$. Different from previous works that use the noisy representation $\tilde{y}$ as an approximation to $\hat{y}$ in distortion, we directly employ $\hat{y}$ with Straight-Through Estimator ~\cite{bengio2013estimating} during optimization. It mainly results from the inconsistency of $\tilde{y}$ in training and $\hat{y}$ in inference, which may negatively influence the generalization of IEM sharing the same parameters $\theta$ during encoding and decoding.

\textbf{Distribution Matching}\quad
This part of training objectives is mainly to enforce the transformation from the y-dependent lost information to a standard Gaussian representation. Denoting the distribution on $z$ that is transformed from the data distribution $q(x)$ as ${f_\theta^z}_\# [q(x)]$, we aim to minimize its difference from the specified y-independent distribution $p(z)$. For practical optimization, we employ the cross-entropy (CE) to measure the difference, which leads to the objective:
{
\begin{equation}
    \begin{aligned}
    	 L_{\mathrm{distribution}}(\theta) &:= \mathrm{CE}( {f_\theta^z}_\# [q(x)], p(z) ) \\ &= - \! \bbE_{{f_\theta^z}_\# [q(x)]} [\log p(z)] \\&= - \bbE_{q(x)} [\log p( z \!=\! f_\theta^z (x) )].
    	\label{eqn:distribution}
    \end{aligned}
\end{equation}
}

The distribution loss encourages $z$ to follow the same target distribution for every $y$, thereby encouraging independence between them.

\textbf{Distillation}\quad
As mentioned in the previous section, we leverage a distillation module to stable our training. The distillation loss based on a teacher model $T_{\delta}$ is defined as:

\begin{align}
	& L_{\mathrm{distillation}}(\theta) := \sum_{n}^{N} \ell_y(f^y_{\theta}(x^{(n)}), T_{\delta}(x^{(n)})),
	\label{eq:distillation}
\end{align}
where $\ell_y$ is a difference metric, and we use $L_2$ loss in practice.

\textbf{Total loss}\quad
Combining these objectives, our total loss for training is:\\
\begin{equation}
    \begin{aligned}
        L_{\mathrm{total}} := \lambda_1 L_{\mathrm{distortion}}+\lambda_2 L_{\mathrm{rate}} & +  \lambda_3 L_{\mathrm{distribution}} \\
        & + \lambda_4 L_{\mathrm{distillation}},
    	\label{eq:total}
    \end{aligned}
\end{equation}
\\
where $\lambda_1$, $\lambda_2$, $\lambda_3$ and $\lambda_4$ are coefficients for balancing different loss terms.

\vspace{-10pt}

\begin{figure} 
    \centering
      \subfloat[]{
       \includegraphics[scale=0.42]{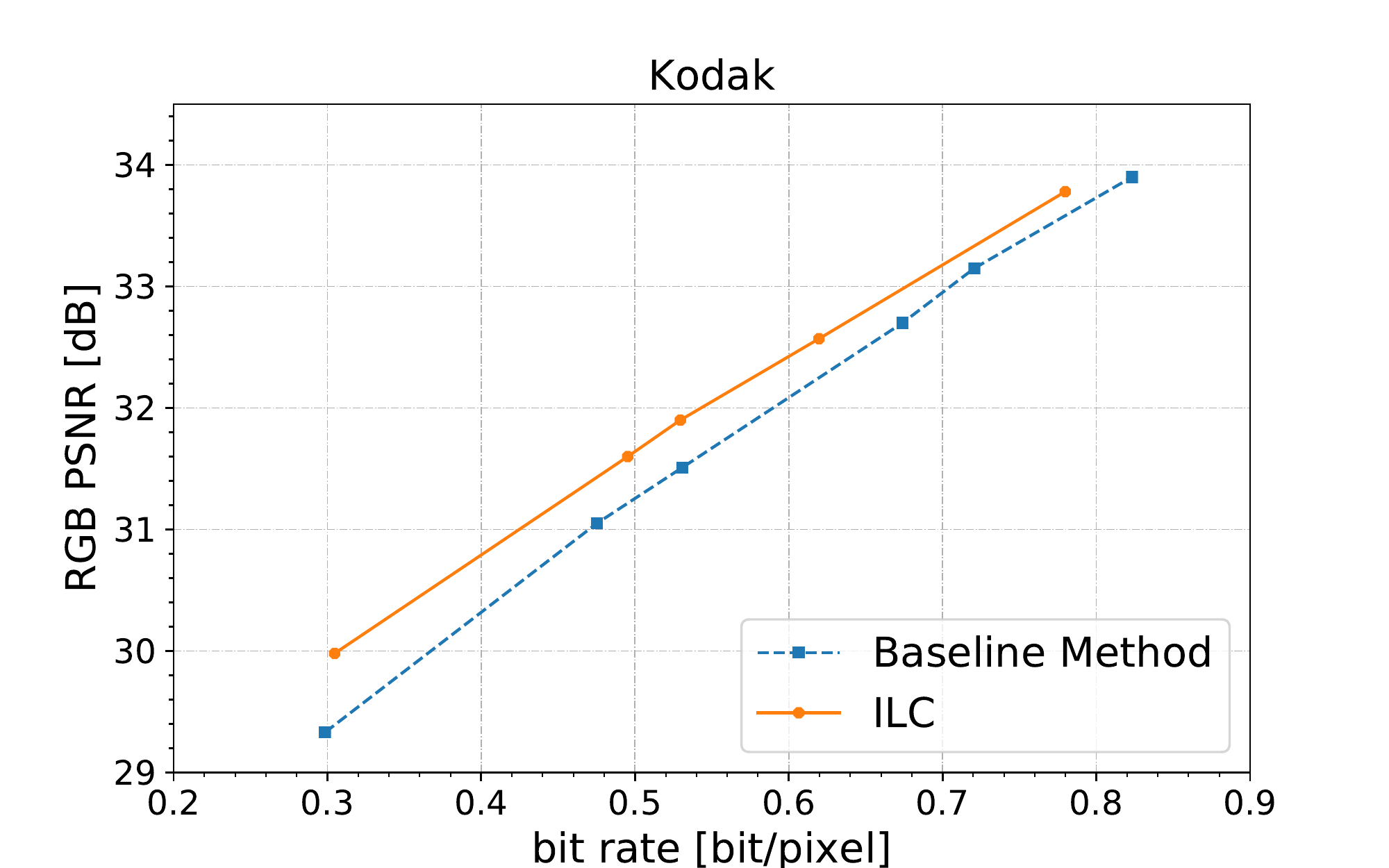}}
    \label{fig.3a}\hfill
      \subfloat[]{
    \includegraphics[scale=0.42]{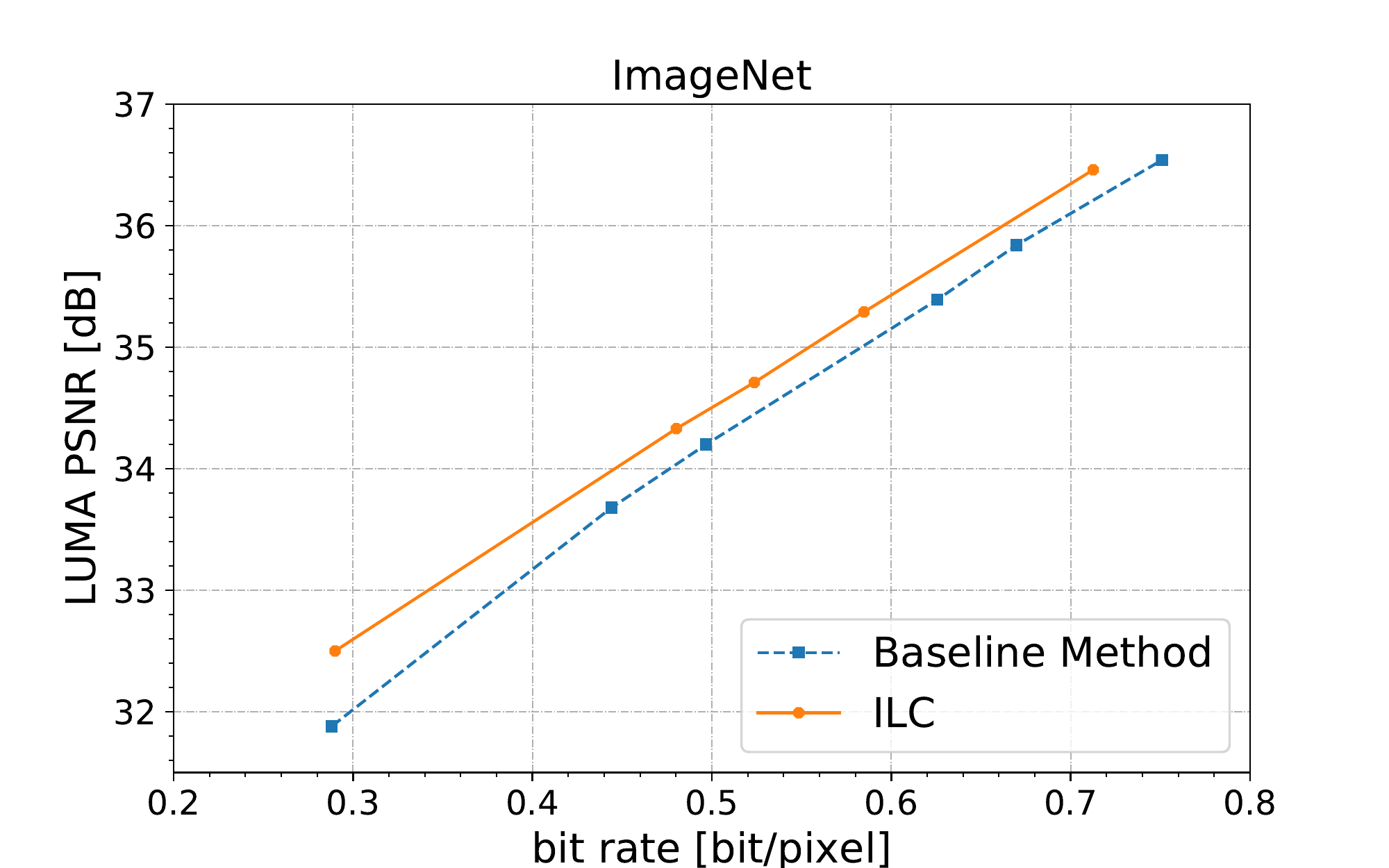}}
        \label{fig.3b}
        \caption{The rate-distortion comparison of the baseline method and ILC distilled from it, where each point is averaged over all the test images in the specified dataset. (a) compares PSNR (RGB) of images from Kodak, and (b) compares PSNR on luma component of 100 images from a randomly sampled test dataset of ImageNet.
        }
    \label{fig.3}
    \vspace{-5pt}
\end{figure}

\vspace{2pt}
\section{Experiments}\label{sec:exp}
\vspace{-2pt}
The dataset we used for training is a subset of ImageNet~\cite{deng2009imagenet}. We filter out 9250 images with a size larger than $512 \times 512$, in order to easily preprocess on data. All the images for training are preprocessed by random rescaling and cropping. Evaluations were performed on the Kodak image dataset~\cite{franzen1999kodak} commonly used as test data for compression problems, and a randomly sampled subset of the ImageNet with 100 uncompressed images, where none of them is used in training.

Our idea is verified on~\cite{balle1}, since it has open-source and reproducible implementation. With the number of filters equal to 256, the teacher encoder is trained following the instructions in ~\cite{balle1}. We use Adam optimization~\cite{kingma2014adam} algorithm for all the parameters with different learning rates on the convolutional auto-encoder and the entropy model, where the former is set to $10^{-4}$, and the other is set to $10^{-3}$. The training and evaluation are performed at each compression rate separately. The compression rate is controlled by adjusting the ratio, i.e. $\lambda$, of distortion and rate in its objective function. To guarantee the training to be thoroughly carried out, we train at least one million iterations and ensure that the performance no longer increases in a certain period of time.

Afterwards, following the architecture in Figure~\ref{fig.1b}, the pre-trained encoder is loaded as our teacher model. We jointly optimize all the parameters in IEM and the entropy model. The training dataset and preprocessing methods are the same as before, except for the learning rate, where we make it gradually decay since 0.1 million iterations. Practically, for different bitrates, we use the same coefficient for distillation, $\lambda_4$, during training, and adjust on all the others, i.e. $\lambda_1$, $\lambda_2$ and $\lambda_3$, in our objective function~\ref{eq:total}.

Aligned with the previous work, we quantified our model with peak noise-signal ratio (PSNR) on both RGB and the luma component of images. We compare the rate-distortion performance of our method to the baseline approach, the teacher model in our framework. For evaluating PSNR, we observe that $z$ has to be sampled from a distribution with lower variance to obtain a better performance. It is consistent with what GLOW ~\cite{kingma2018glow} does, which samples from a distribution with shrinked variance to prevent mode collapse on generated images. In our case, the priority is not variance but the performance in reconstruction, so instead of sampling, we take the most-likely $z$ from the specified distribution. Compared with the baseline method, with the same compression ratio, ILC improves the PSNR performance by around 0.4dB. And empirically, as the compression ratio rises (i.e. bitrate decreases), ILC can better boost the performance. Intuitively, under more severe demands on bitrate, the convolutional auto-encoder tends to undertake more responsibility in further extracting the refined representation of data, which leads to more irreversible loss of information. In this case, explicitly modelling the high-frequency signals turns into the key to success.

To further verify that ILC's superiority over auto-encoder brings the improvement, we use this result to compare with ILC guided by an early-stopped teacher model. It shows a strong positive correlation between the performance of ILC and its teacher model, which testifies our thoughts that ILC is strongly guided and would always look for a manifold surrounding it. ILC's preponderance is mainly in the reconstruction, so it could be leveraged to improve the performance jointly with other more advanced entropy models.

\vspace{-4pt}
\section{Conclusion}
\vspace{-2pt}
In this paper, we propose a novel framework ILC for lossy image compression, with explicitly modelling the information loss during encoding. By doing this, the ill-posed problem of the post-reconstruction stage is largely mitigated. To achieve our purpose, we design an invertible encoding module to replace the encoder-decoder network in previous methods, which is obviously more suitable for this pair of inverse tasks. With the latent variable's statistical knowledge, IEM can reconstruct the original input image from the compressed bit-stream with high quality by drawing a sample from a pre-specific distribution. To overcome the intractable challenges in optimization, we propose a knowledge distillation module to provide the soft labels from a teacher model, which significantly accelerates the training procedure. Extensive experiments demonstrate that our framework ILC significantly improves performance based on existing methods.

\clearpage

\bibliographystyle{IEEEtran}
\bibliography{ICN.bib}

\newpage

\textbf{\Large Appendix} \vspace{20pt}
\section{Experimental Details}
\textbf{Network Structure} \quad The coupling layers are parametrized using neural networks with a bottleneck-like structure, which consists of three convolutional layers, where the first and last is $K \times K$, and the middle one is $1 \times 1$. This is designed to improve efficiency in using parameters by allowing a larger number of channels under an acceptable computing complexity. In our implementation, the detail of architecture is specified in Table~\ref{tab:network structure}.
\vspace{2pt}
\begin{table}[ht]
    \centering
    \begin{tabular}{c|c|c|c}
        \hline
    	Input Size & Input Channel  &  Width & Kernel Size (K)\\
    	\hline
    	\hline
        64 $\times$ 64 & 48 & 128 & 5 $\times$ 5 \\
        \hline
        32 $\times$ 32 & 192 & 256 & 3 $\times$ 3 \\
        \hline
        16 $\times$ 16 & 768 & 1024 & 3 $\times$ 3
    \end{tabular}
    \caption{Architectures for IEM}
    \label{tab:network structure}
    \vspace{-10pt}
\end{table}
\\
\textbf{Training Details} \quad
Throughout training, we use Adam optimizer~\cite{kingma2014adam} with $\beta_1=0.9$ and $\beta_2=0.999$. For the learning rate, different initial values for IEM and the entropy model are used, denoted as $lr^{init}_{IEM}$ and $lr^{init}_{EM}$ respectively. We let both of them decay from the 0.1 million iterations, which is computed as 
\begin{align}
	& lr^{iter=i}_{*} = lr^{init}_{*} \times max\{\gamma^{i}, 1\},
	\label{eq:lr}
\end{align}
\\
, where $* \in \{IEM, EM\}$, $i$ is a step counter, and $\gamma$ is a decay factor that is practically around 0.999995. $lr^{init}_{IEM}$ and $lr^{init}_{EM}$ are set to $10^{-4}$ and $10^{-3}$ respectively.

As for the training objective, the coefficients need to be carefully tuned for each bitrate. Empirically, we set $\lambda_2 = 1$ and $\lambda_4 = 10^{-2}$, and use grid search to look for optimal weights of $\lambda_1$ and $\lambda_3$ ranging from $10^{-1}$ to $10^1$. Notice, it does not mean $\lambda_4$ is optimal in all the scenarios, but experimentally there is no significant difference between the change of performance when simultaneously changing $\lambda_2$ and $\lambda_4$, and adjusting $\lambda_2$ only.\\\\
\vspace{0pt}
\textbf{Ramdomly Sampled Dataset for Evaluation} \quad
The subset of ImageNet\cite{deng2009imagenet} we used for evaluation is randomly sampled, which is listed in table \ref{tab:subset}.
\begin{table}[!t]
    \centering
    \vspace{-50pt}
    \begin{tabular}{ c c c }
    \hline
     \textit{\textbf{ILSVRC2012\_test\_00001444}} & \textit{\textbf{ILSVRC2012\_test\_00005577}} \\ \textit{\textbf{ILSVRC2012\_test\_00006189}} &  
     \textit{\textbf{ILSVRC2012\_test\_00007775}} \\ \textit{\textbf{ILSVRC2012\_test\_00008278}} & \textit{\textbf{ILSVRC2012\_test\_00008440}} \\
     \textit{\textbf{ILSVRC2012\_test\_00008460}} & \textit{\textbf{ILSVRC2012\_test\_00009563}} \\ \textit{\textbf{ILSVRC2012\_test\_00010174}} &
     \textit{\textbf{ILSVRC2012\_test\_00011618}} \\ \textit{\textbf{ILSVRC2012\_test\_00012189}} & \textit{\textbf{ILSVRC2012\_test\_00012215}} \\
     \textit{\textbf{ILSVRC2012\_test\_00012448}} & \textit{\textbf{ILSVRC2012\_test\_00012492}} \\ \textit{\textbf{ILSVRC2012\_test\_00012706}} &
     \textit{\textbf{ILSVRC2012\_test\_00012814}} \\ \textit{\textbf{ILSVRC2012\_test\_00013452}} & \textit{\textbf{ILSVRC2012\_test\_00016082}} \\
     \textit{\textbf{ILSVRC2012\_test\_00020129}} & \textit{\textbf{ILSVRC2012\_test\_00023175}} \\ \textit{\textbf{ILSVRC2012\_test\_00023432}} &
     \textit{\textbf{ILSVRC2012\_test\_00023806}} \\ \textit{\textbf{ILSVRC2012\_test\_00024649}} & \textit{\textbf{ILSVRC2012\_test\_00027574}} \\
     \textit{\textbf{ILSVRC2012\_test\_00028721}} & \textit{\textbf{ILSVRC2012\_test\_00029016}} \\ \textit{\textbf{ILSVRC2012\_test\_00029428}} &
     \textit{\textbf{ILSVRC2012\_test\_00030152}} \\ \textit{\textbf{ILSVRC2012\_test\_00030240}} & \textit{\textbf{ILSVRC2012\_test\_00030535}} \\
     \textit{\textbf{ILSVRC2012\_test\_00030594}} & \textit{\textbf{ILSVRC2012\_test\_00033552}} \\ \textit{\textbf{ILSVRC2012\_test\_00035652}} &
     \textit{\textbf{ILSVRC2012\_test\_00036628}} \\ \textit{\textbf{ILSVRC2012\_test\_00037322}} & \textit{\textbf{ILSVRC2012\_test\_00037832}} \\
     \textit{\textbf{ILSVRC2012\_test\_00038678}} & \textit{\textbf{ILSVRC2012\_test\_00038827}} \\ \textit{\textbf{ILSVRC2012\_test\_00039565}} &
     \textit{\textbf{ILSVRC2012\_test\_00040346}} \\ \textit{\textbf{ILSVRC2012\_test\_00040449}} & \textit{\textbf{ILSVRC2012\_test\_00042420}} \\
     \textit{\textbf{ILSVRC2012\_test\_00042491}} & \textit{\textbf{ILSVRC2012\_test\_00042549}} \\ \textit{\textbf{ILSVRC2012\_test\_00042583}} &
     \textit{\textbf{ILSVRC2012\_test\_00042631}} \\ \textit{\textbf{ILSVRC2012\_test\_00043029}} & \textit{\textbf{ILSVRC2012\_test\_00043873}} \\
     \textit{\textbf{ILSVRC2012\_test\_00046159}} & \textit{\textbf{ILSVRC2012\_test\_00047576}} \\ \textit{\textbf{ILSVRC2012\_test\_00049924}} &
     \textit{\textbf{ILSVRC2012\_test\_00050181}} \\ \textit{\textbf{ILSVRC2012\_test\_00051384}} & \textit{\textbf{ILSVRC2012\_test\_00053071}} \\
     \textit{\textbf{ILSVRC2012\_test\_00053603}} & \textit{\textbf{ILSVRC2012\_test\_00054755}} \\ \textit{\textbf{ILSVRC2012\_test\_00055533}} &
     \textit{\textbf{ILSVRC2012\_test\_00063855}} \\ \textit{\textbf{ILSVRC2012\_test\_00063995}} & \textit{\textbf{ILSVRC2012\_test\_00066697}} \\
     \textit{\textbf{ILSVRC2012\_test\_00066914}} & \textit{\textbf{ILSVRC2012\_test\_00068457}} \\ \textit{\textbf{ILSVRC2012\_test\_00068596}} &
     \textit{\textbf{ILSVRC2012\_test\_00070325}} \\ \textit{\textbf{ILSVRC2012\_test\_00071340}} & \textit{\textbf{ILSVRC2012\_test\_00071954}} \\
     \textit{\textbf{ILSVRC2012\_test\_00073228}} & \textit{\textbf{ILSVRC2012\_test\_00073412}} \\ \textit{\textbf{ILSVRC2012\_test\_00074198}} & 
     \textit{\textbf{ILSVRC2012\_test\_00075878}} \\ \textit{\textbf{ILSVRC2012\_test\_00076257}} & \textit{\textbf{ILSVRC2012\_test\_00078714}} \\
     \textit{\textbf{ILSVRC2012\_test\_00078897}} & \textit{\textbf{ILSVRC2012\_test\_00080599}} \\ \textit{\textbf{ILSVRC2012\_test\_00081961}} &
     \textit{\textbf{ILSVRC2012\_test\_00082349}} \\ \textit{\textbf{ILSVRC2012\_test\_00085480}} & \textit{\textbf{ILSVRC2012\_test\_00085775}} \\
     \textit{\textbf{ILSVRC2012\_test\_00086081}} & \textit{\textbf{ILSVRC2012\_test\_00086280}} \\ \textit{\textbf{ILSVRC2012\_test\_00086391}} &
     \textit{\textbf{ILSVRC2012\_test\_00086845}} \\ \textit{\textbf{ILSVRC2012\_test\_00087796}} & \textit{\textbf{ILSVRC2012\_test\_00087939}} \\
     \textit{\textbf{ILSVRC2012\_test\_00089036}} & \textit{\textbf{ILSVRC2012\_test\_00089377}} \\ \textit{\textbf{ILSVRC2012\_test\_00089680}} &
     \textit{\textbf{ILSVRC2012\_test\_00090568}} \\ \textit{\textbf{ILSVRC2012\_test\_00090956}} & \textit{\textbf{ILSVRC2012\_test\_00093267}} \\
     \textit{\textbf{ILSVRC2012\_test\_00093343}} & \textit{\textbf{ILSVRC2012\_test\_00096391}} \\ \textit{\textbf{ILSVRC2012\_test\_00096900}} &
     \textit{\textbf{ILSVRC2012\_test\_00099475}} \\ \textit{\textbf{ILSVRC2012\_test\_00099912}} & \textit{\textbf{ILSVRC2015\_test\_00002901}} \\
     \textit{\textbf{ILSVRC2015\_test\_00010740}} & & \\
     \hline
    \end{tabular}
    \caption{All the file names in the subset of ImageNet used for evaluation.}
    \label{tab:subset}
    \vspace{30pt}
\end{table}
\clearpage
\vspace{25pt}
\section{Performance without Quantization}
Shown in table \ref{tab:wq}, We test our model without quantization and find that its performance is remarkably improved. Contrarily, with an irreversible auto-encoder, there is only a slight performance increment, which shows the preponderance of invertibility in the process of reconstruction. Such a huge gap result from quantization implies a tremendous potential in information modelling, which points out a direction for prospective works. 
\begin{table}[!h]
    \centering
    \begin{tabular}{c|c|c|c}
    \multicolumn{2}{c|}{Baseline} & \multicolumn{2}{c}{ILC}\\
    \hline
    NQ (dB) / Q (dB) & bpp & NQ (dB) / Q (dB) & bpp\\
    \hline
    31.61 / 29.84 & 0.2984 & 40.62 / 30.59 & 0.3047 \\
    \hline
    33.45 / 31.63 & 0.4752 & 42.43 / 32.36 & 0.4917 \\
    \hline
    34.05 / 32.24 & 0.5308 & 43.07 / 32.59 & 0.5295 \\
    \hline
    35.22 / 33.42 & 0.674 & 44.11 / 33.4 & 0.6197 \\
    \hline
    36.36 / 34.61 & 0.8232 & 45.49 / 34.62 & 0.7798 \\
    \end{tabular}
    \caption{We denote the evaluation performance with quantization as Q, and the one without quantization as NQ. The performance in ILC soars more significantly as quantization is removed.}
    \label{tab:wq}
\end{table}
\section{Ablation Study}
\begin{figure}[!h]
    \centering
      \subfloat[]{
       \includegraphics[scale=0.4]{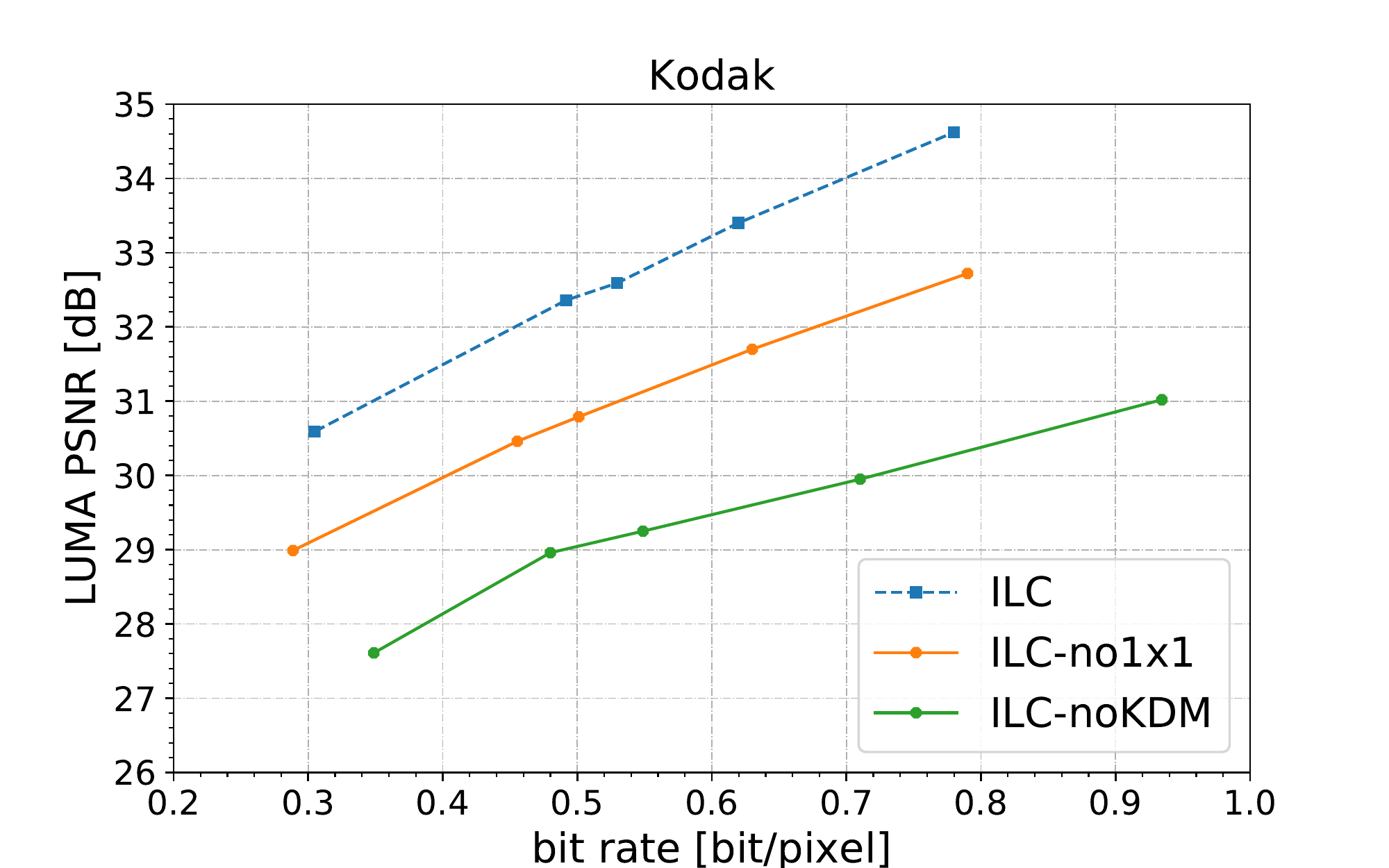}}
    \label{fig.12a}
      \caption{To further verify the efficiency of our framework, we conduct ablation study on several modules. The suffix \textit{-no1$\times$1} means removing all the invertible $1 \times 1$ convolutions. The suffix \textit{-noKDM} means training without knowledge distillation, i.e. specifically speaking, there are only three components in the objective, corresponding to distortion, rate, and distribution matching.}
      \label{fig.12} 
\end{figure}
\section{More Results}
\clearpage
\begin{figure*}
    \centering
      \subfloat[]{
       \includegraphics[scale=0.41]{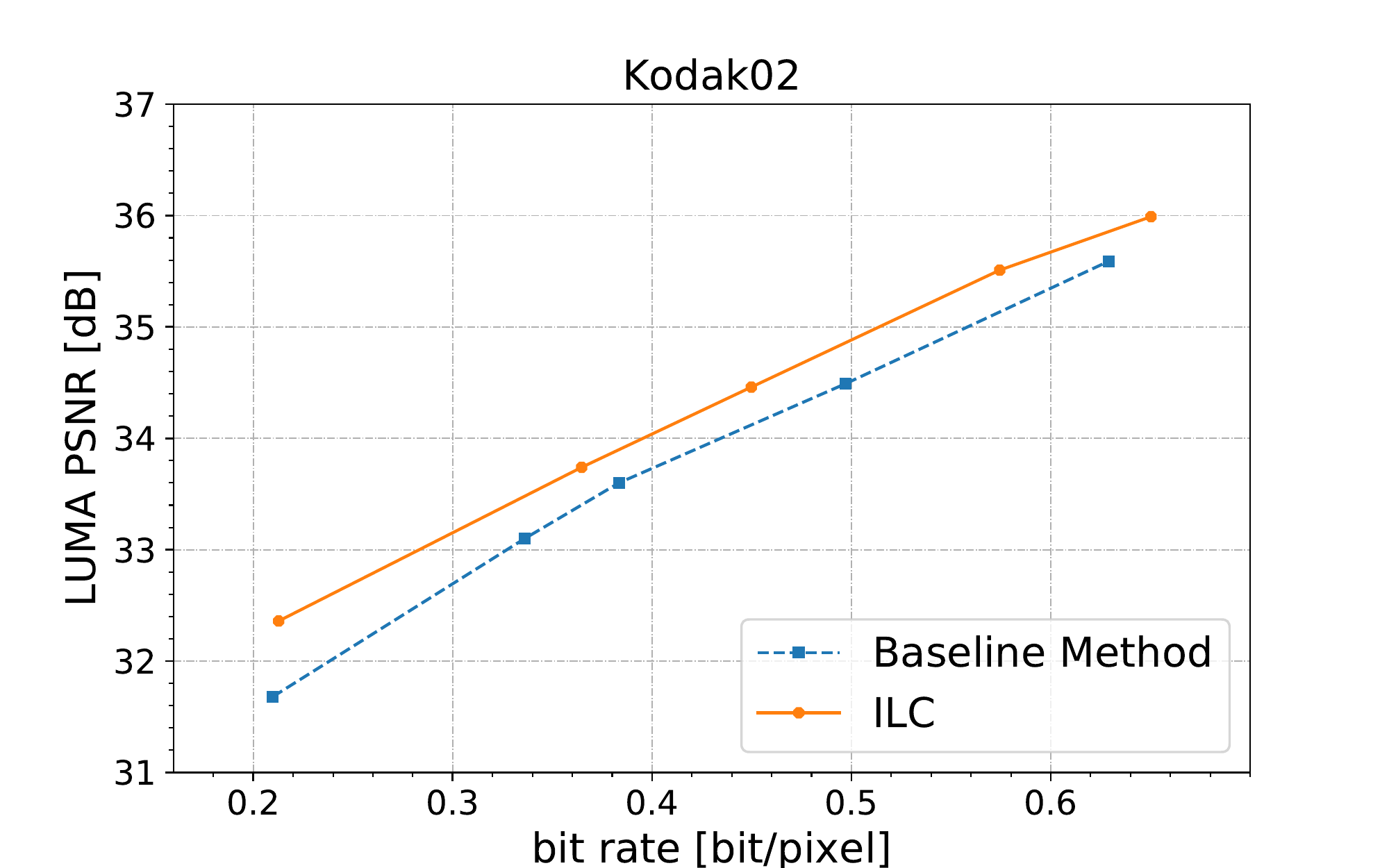}}
    \label{fig.4a}\hfill
      \subfloat[]{
        \includegraphics[scale=0.41]{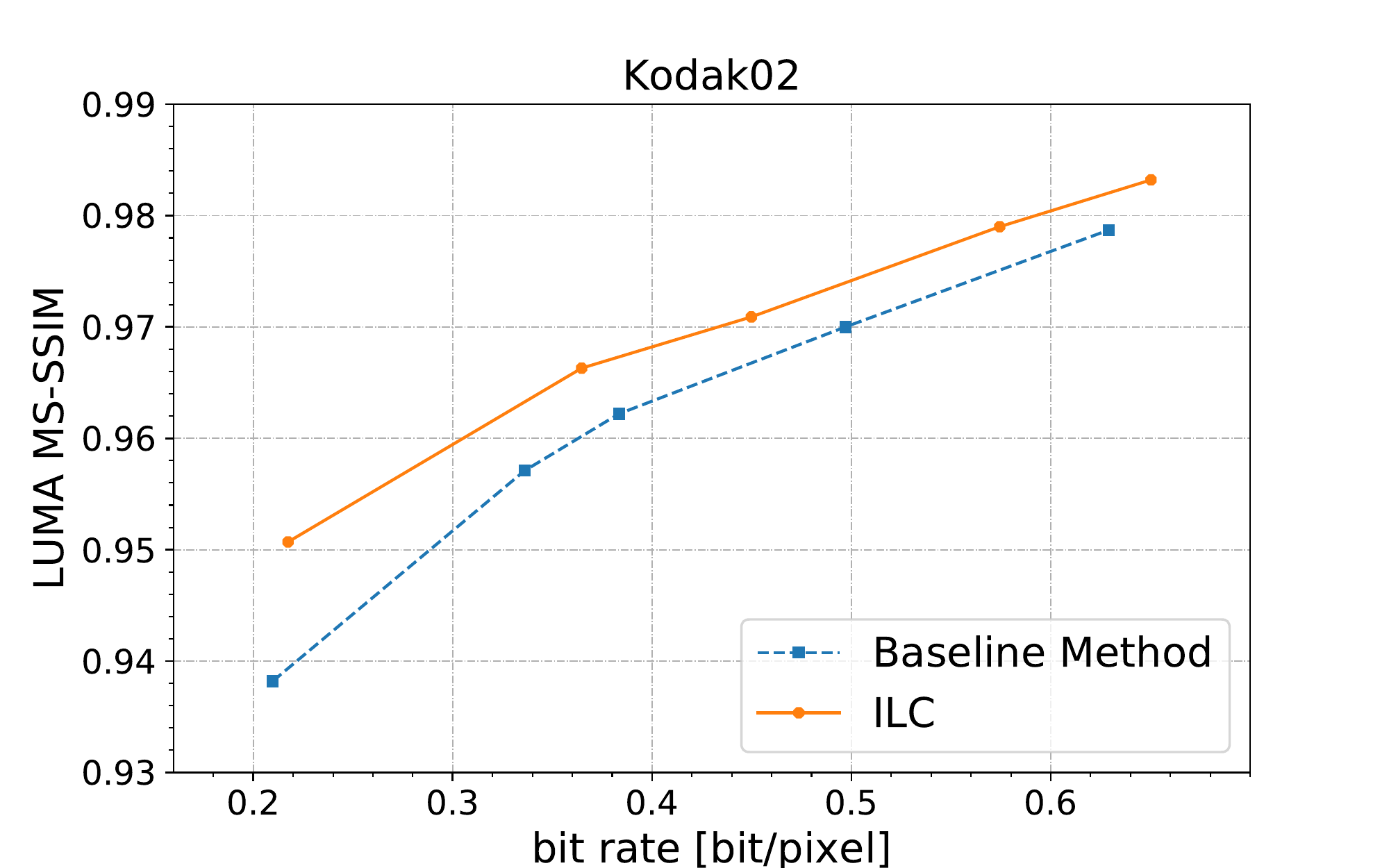}}
    \label{fig.4b}
      \caption{R-D curve on Kodak-02 compared between our framework and baseline method. (a) illustrates the performance on (luma) PSNR, and (b) illustrates the performance on (luma) MS-SSIM}
      \label{fig.4} 
\end{figure*}

\begin{figure*} 
    \centering
      \subfloat[]{
       \includegraphics[scale=0.2]{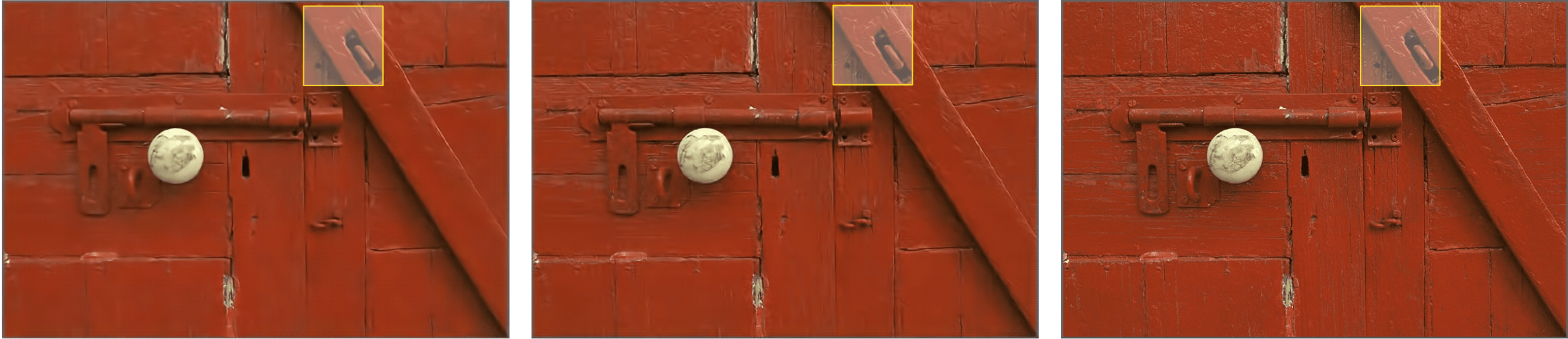}}
    \label{fig.5a}\hfill
      \subfloat[]%
      [Baseline \par(luma) PSNR: 31.68 dB, \par(rgb) PSNR: 30.62 dB, \par(luma) MS-SSIM: 0.9382, \par bpp: 0.2127]%
      {\includegraphics[scale=1]{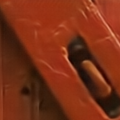}}
    \label{fig.5b}
        \subfloat[]%
        [ILC \par (luma) PSNR: \textbf{32.26 dB}, \par (rgb) PSNR: \textbf{31.17 dB}, \par (luma) MS-SSIM: \textbf{0.9507}, \par bpp: 0.2174]%
        {\includegraphics[scale=1]{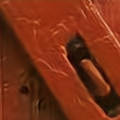}}
    \label{fig.5c}
        \subfloat[Original Image]{
        \includegraphics[scale=1]{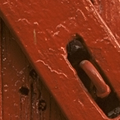}}
    \label{fig.5d}
      \caption{The performance comparison on Kodak-02. (a) shows three images, which are the reconstruction from baseline, the reconstruction from ILC and the original image, respectively from left to right. (b), (c) and (d) are image patches from (a)-left, (a)-middle, and (a)-right respectively.}
      \label{fig.5} 
\end{figure*}

\begin{figure*} 
    \centering
      \subfloat[]{
       \includegraphics[scale=0.41]{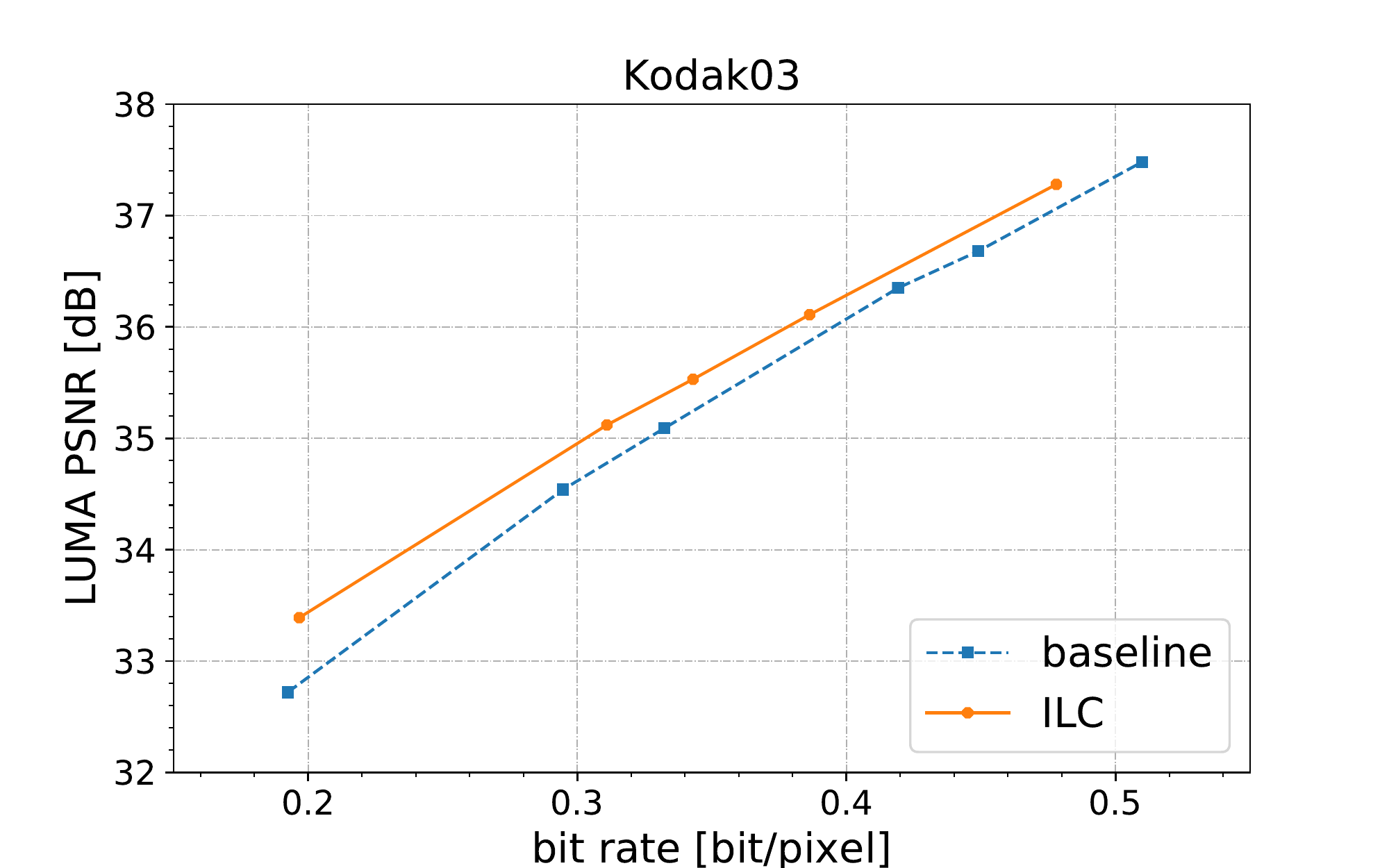}}
    \label{fig.6a}\hfill
      \subfloat[]{
        \includegraphics[scale=0.41]{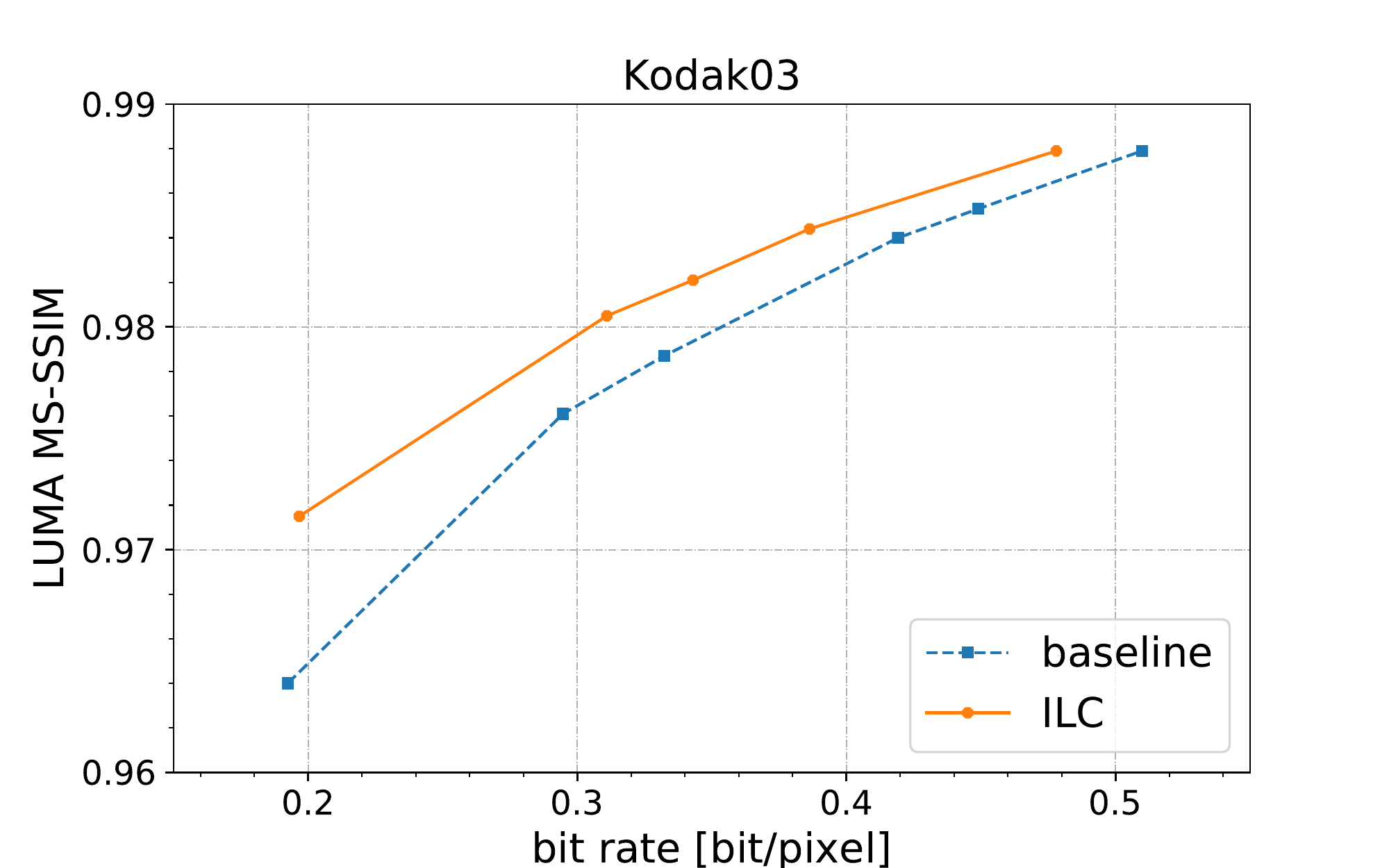}}
    \label{fig.6b}
      \caption{R-D curve on Kodak-03 compared between our framework and baseline method. (a) illustrates the performance on (luma) PSNR, and (b) illustrates the performance on (luma) MS-SSIM}
      \label{fig.6} 
\end{figure*}
\begin{figure*} 
    \centering
      \subfloat[]{
       \includegraphics[scale=0.2]{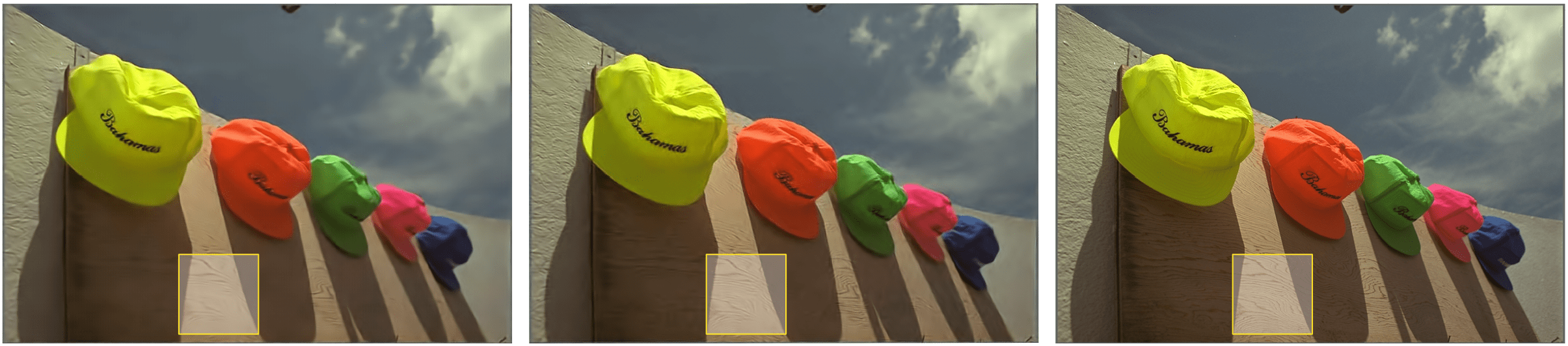}}
    \label{fig.7a}\hfill
      \subfloat[]%
      [Baseline \par (luma) PSNR: 32.72 dB, \par (rgb) PSNR: 32.20 dB, \par (luma) MS-SSIM: 0.9640, \par bpp: 0.1924]%
      {\includegraphics[scale=1]{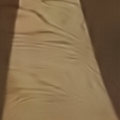}}
    \label{fig.7b}
        \subfloat[]%
        [ILC \par (luma) PSNR: \textbf{33.39 dB}, \par (rgb) PSNR: \textbf{32.72 dB}, \par (luma) MS-SSIM: \textbf{0.9715}, \par bpp: 0.1967]%
        {\includegraphics[scale=1]{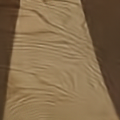}}
    \label{fig.7c}
        \subfloat[Original Image]{
        \includegraphics[scale=1]{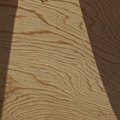}}
    \label{fig.7d}
      \caption{The performance comparison on Kodak-03. (a) shows three images, which are the reconstruction from baseline, the reconstruction from ILC and the original image, respectively from left to right. (b), (c) and (d) are image patches from (a)-left, (a)-middle, and (a)-right respectively.}
      \label{fig.7} 
\end{figure*}

\begin{figure*} 
    \centering
      \subfloat[]{
       \includegraphics[scale=0.41]{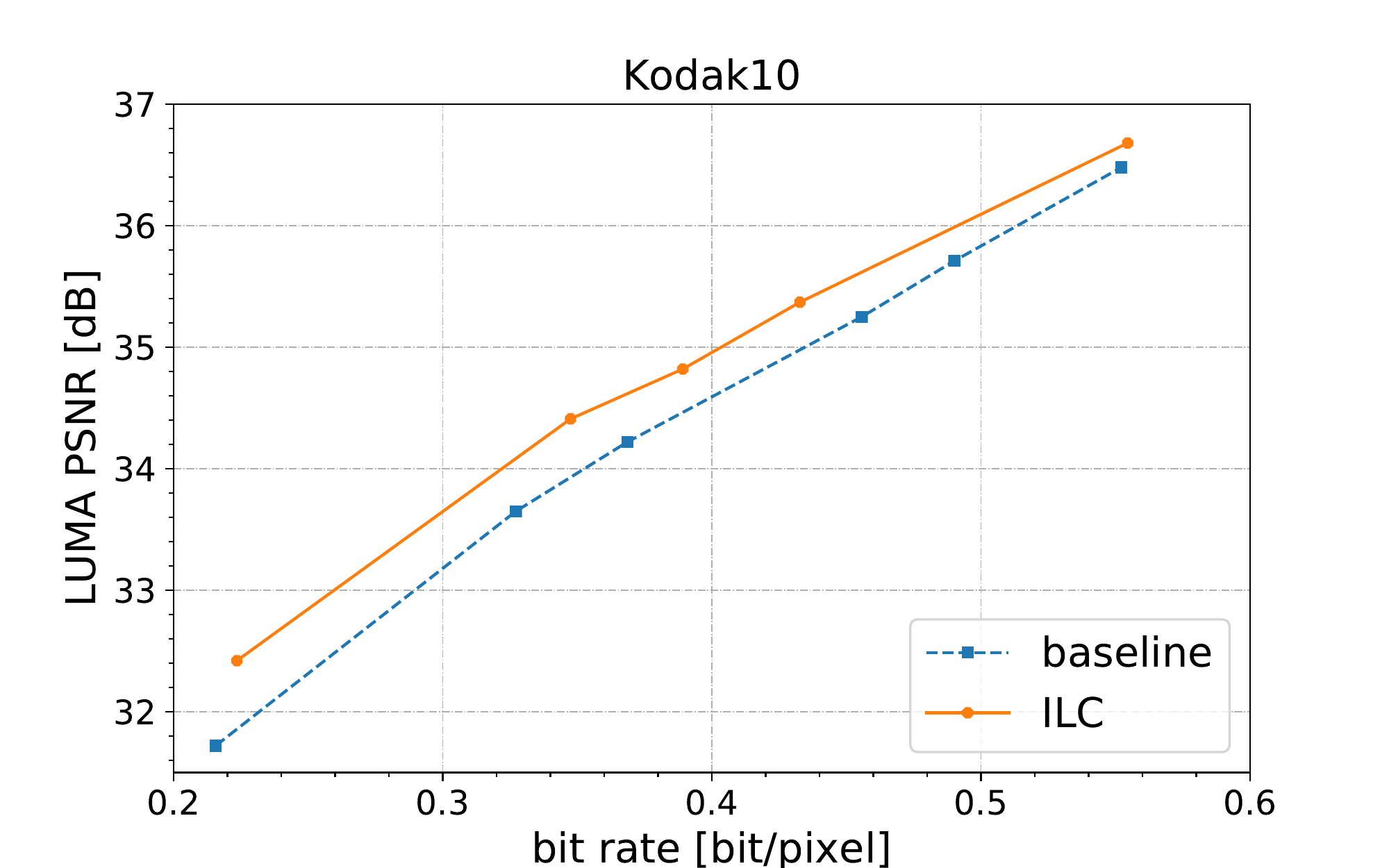}}
    \label{fig.8a}\hfill
      \subfloat[]{
        \includegraphics[scale=0.41]{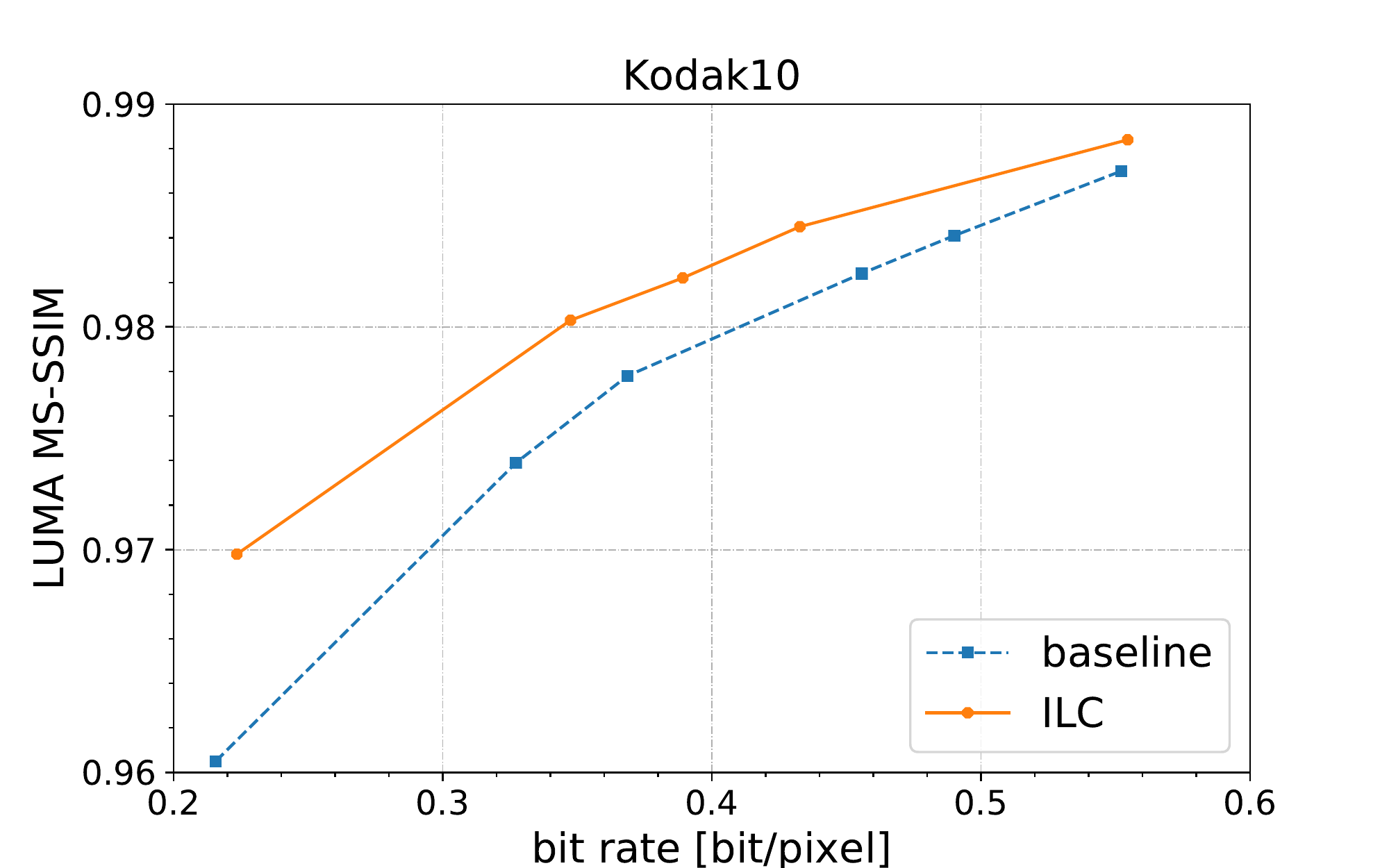}}
    \label{fig.8b}
      \caption{R-D curve on Kodak-10 compared between our framework and baseline method. (a) illustrates the performance on (luma) PSNR, and (b) illustrates the performance on (luma) MS-SSIM}
      \label{fig.8} 
\end{figure*}
\begin{figure*} 
    \centering
      \subfloat[]{
       \includegraphics[scale=0.2]{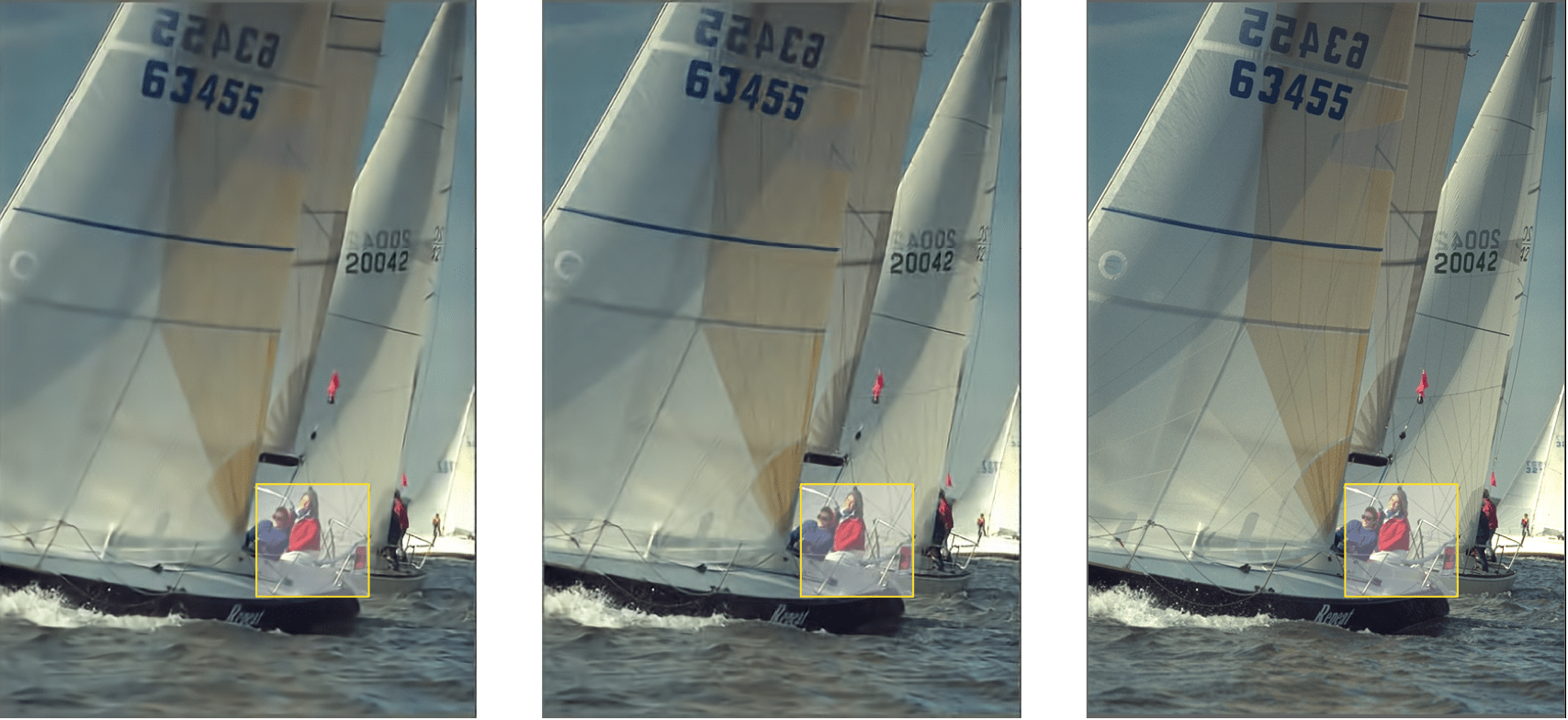}}
    \label{fig.9a}\\
      \subfloat[]%
      [Baseline \par (luma) PSNR: 31.72 dB, \par (rgb) PSNR: 30.91 dB, \par (luma) MS-SSIM: 0.9605, \par bpp: 0.2156]%
      {\includegraphics[scale=1]{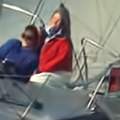}}
    \label{fig.9b}
        \subfloat[]%
        [ILC \par (luma) PSNR: \textbf{32.42 dB}, \par (rgb) PSNR: \textbf{31.50 dB}, \par (luma) MS-SSIM: \textbf{0.9698}, \par bpp: 0.2235]%
        {\includegraphics[scale=1]{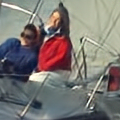}}
    \label{fig.9c}
        \subfloat[Original Image]{
        \includegraphics[scale=1]{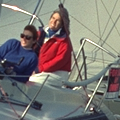}}
    \label{fig.9d}
      \caption{The performance comparison on Kodak-10. (a) shows three images, which are the reconstruction from baseline, the reconstruction from ILC and the original image, respectively from left to right. (b), (c) and (d) are image patches from (a)-left, (a)-middle, and (a)-right respectively.}
      \label{fig.9} 
\end{figure*}

\begin{figure*} 
    \centering
      \subfloat[]{
       \includegraphics[scale=0.41]{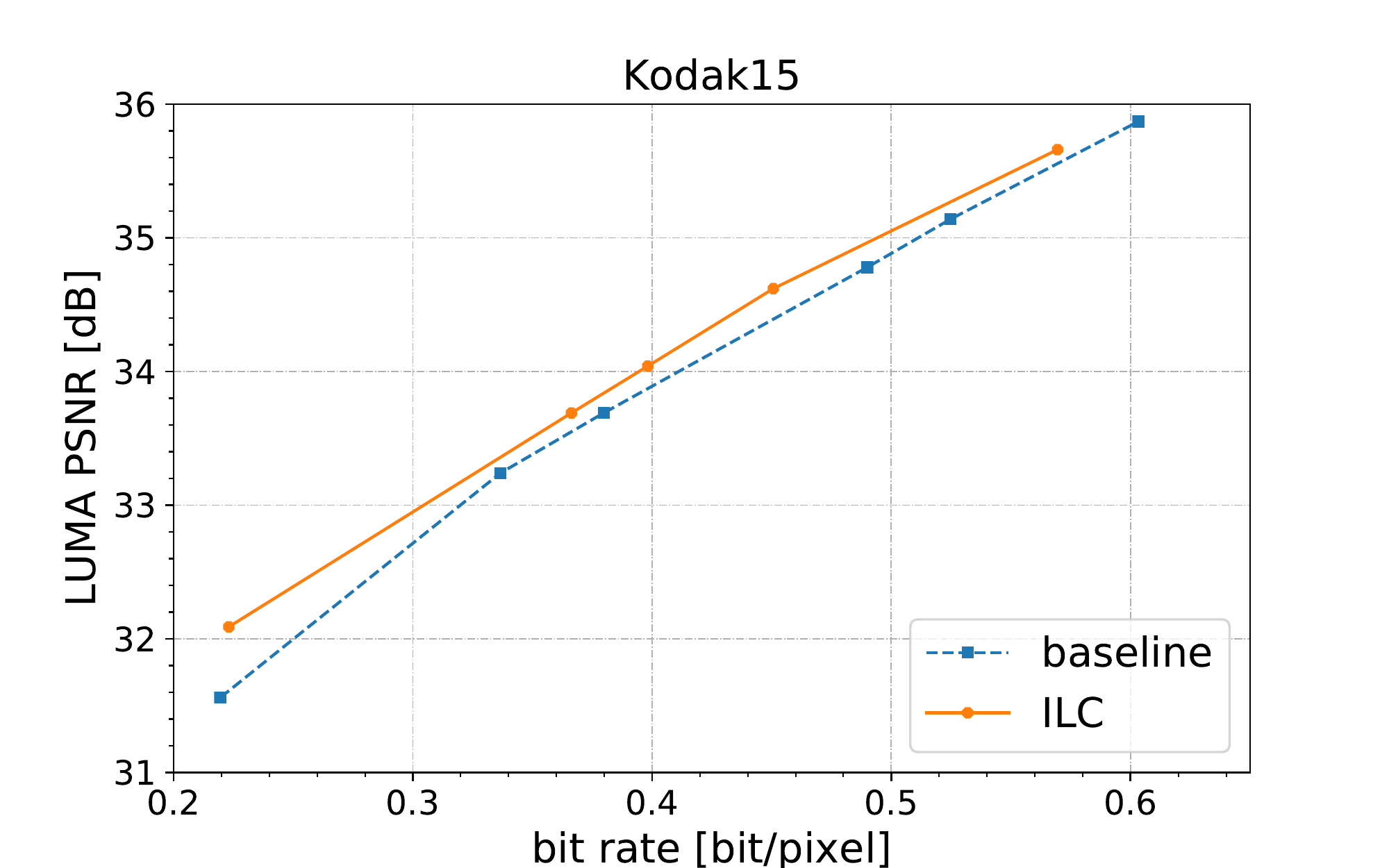}}
    \label{fig.10a}\hfill
      \subfloat[]{
        \includegraphics[scale=0.41]{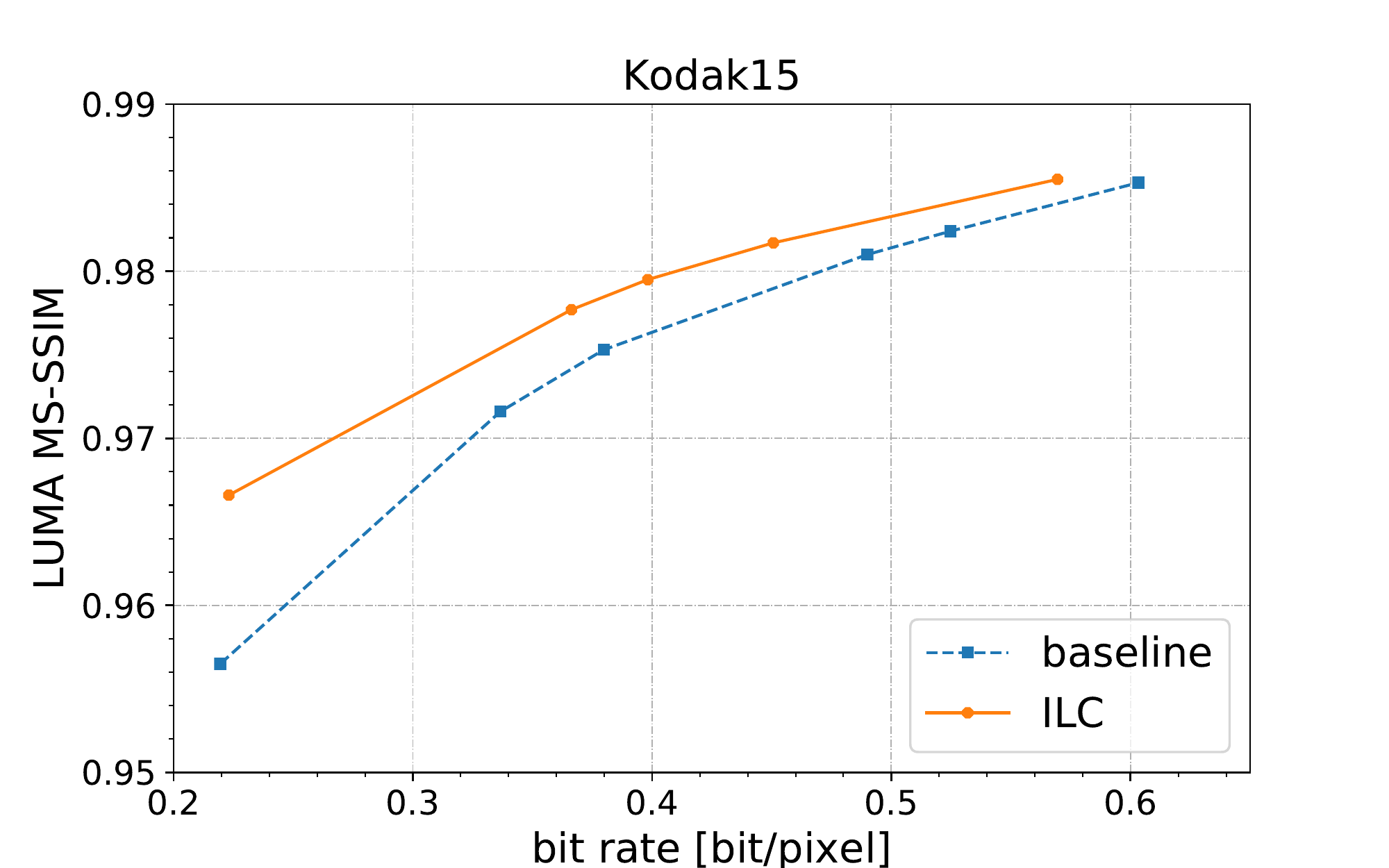}}
    \label{fig.10b}
      \caption{R-D curve on Kodak-15 compared between our framework and baseline method. (a) illustrates the performance on (luma) PSNR, and (b) illustrates the performance on (luma) MS-SSIM}
      \label{fig.10} 
\end{figure*}
\begin{figure*} 
    \centering
      \subfloat[]{
       \includegraphics[scale=0.2]{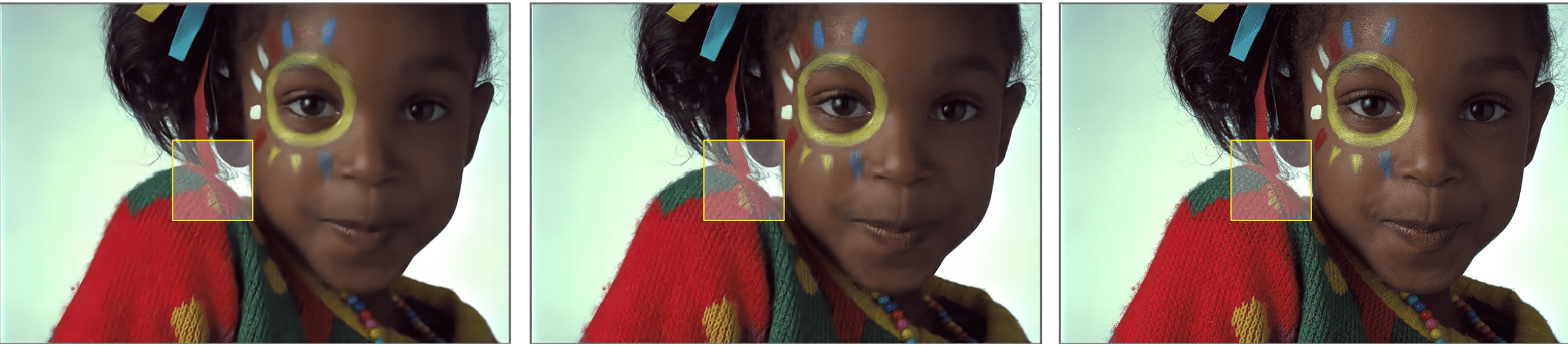}}
    \label{fig.11a}\\
      \subfloat[]%
      [Baseline \par (luma) PSNR: 31.56 dB, \par (rgb) PSNR: 30.75 dB, \par (luma) MS-SSIM: 0.9565, \par bpp: 0.2195]%
      {\includegraphics[scale=1]{recons-kodak-10.png}}
    \label{fig.11b}
        \subfloat[]%
        [ILC \par (luma) PSNR: \textbf{32.09 dB}, \par (rgb) PSNR: \textbf{31.19 dB}, \par (luma) MS-SSIM: \textbf{0.9666}, \par bpp: 0.2231]%
        {\includegraphics[scale=1]{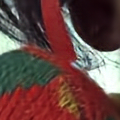}}
    \label{fig.11c}
        \subfloat[Original Image]{
        \includegraphics[scale=1]{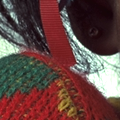}}
    \label{fig.11d}
      \caption{The performance comparison on Kodak-15. (a) shows three images, which are the reconstruction from baseline, the reconstruction from ILC and the original image, respectively from left to right. (b), (c) and (d) are image patches from (a)-left, (a)-middle, and (a)-right respectively.}
      \label{fig.11} 
\end{figure*}

\clearpage

\end{document}